**Title:** Lifetime-configurable soft robots via photodegradable silicone elastomer composites.

**Authors:** Min-Ha Oh,[1]† Young-Hwan Kim,[1]† Seung-Min Lee,[1]† Gyeong-Seok Hwang,[3] Kyung-Sub Kim,[1] Jae-Young Bae,[1] Ju-Young Kim,[3] Ju-Yong Lee,[1] Yu-Chan Kim,[4,5] Sang Yup Kim,[6]* Seung-Kyun Kang[1,2]*

**Affiliations:**

[1] Department of Materials Science and Engineering, Seoul National University; 1 Gwanak-ro, Gwanak-gu, Seoul, 08826 Republic of Korea

[2] Research Institute of Advanced Materials (RIAM), Seoul National University; 1 Gwanak-ro, Gwanak-gu, Seoul, 08826 Republic of Korea.

[3] Department of Materials Science and Engineering, UNIST (Ulsan National Institute of Science and Technology); Ulsan, 44919, Republic of Korea

[4] Center for Biomaterials, Biomedical Research Institute, Korea Institute of Science and Technology (KIST), Seoul, 02792, Republic of Korea

[5] Division of Bio-Medical Science and Technology, KIST School, Korea University of Science and Technology, Seoul, 02792, Republic of Korea

[6] Department of Mechanical Engineering, Sogang University; 35 Baekbeom-ro, Mapo-gu, Seoul, 04107 Republic of Korea

*Corresponding author. Email: Seung-Kyun Kang (S.-K. Kang), kskg7227@snu.ac.kr; Sang Yup Kim (S.Y. Kim) sangyupkim@sogang.ac.kr.

†These authors contributed equally to this work

**Abstract:** Developing soft robots that can control their own life-cycle and degrade on-demand while maintaining hyper-elasticity is a significant research challenge. On-demand degradable soft robots, which conserve their original functionality during operation and rapidly degrade under specific external stimulation, present the opportunity to self-direct the disappearance of temporary robots. This study proposes soft robots and materials that exhibit excellent mechanical stretchability and can degrade under ultraviolet (UV) light by mixing a fluoride-generating diphenyliodonium hexafluorophosphate (DPI-HFP) with a silicone resin. Spectroscopic analysis revealed the mechanism of Si-O-Si backbone cleavage using fluoride ion ($F^-$), which was generated from UV exposed DPI-HFP. Furthermore, photo-differential scanning calorimetry (DSC) based thermal analysis indicated increased decomposition kinetics at increased temperatures. Additionally, we demonstrated a robotics application of this composite by fabricating a gaiting robot. The integration of soft electronics, including strain sensors, temperature sensors, and photodetectors, expanded the robotic functionalities. This study provides a simple yet novel strategy for designing lifecycle mimicking soft robotics that can be applied to reduce soft robotics waste, explore hazardous areas where retrieval of robots is impossible, and ensure hardware security with on-demand destructive material platforms.

**One-Sentence Summary:** We propose on-demand transient soft robotic materials and system whose degradation is triggered by UV exposure.

**Main Text:**

**INTRODUCTION**

Soft robotic materials with hyper-elasticity and tunable functionalities have been an integral part of soft robotics (*1,2*). Unlike conventional robots with limited degrees of freedom, soft robots demonstrate advanced functional adaptability and enable convoluted movements, such as delicate handling of vulnerable objects or adapting to uncertain environments (*3,4*). The rapid evolution of soft robots improved their capability of mimicking the individual functions of biological systems as well as emulating the lifecycle of living organisms, such as growth (*5,6*), healing (*7,8*), transition (*9*), transformation (*10*), and death (*11*). Considering the lifecycle mimicry by robotic systems, the "death" and "disposal" of soft robots after their operational lifetime have gained wide attention in terms of creating a sustainable environment for future robotic systems (*12*). Transient soft robots, whose components degrade or dissolve in a controlled manner, are emerging as a solution for the effective death management of soft robots that are no longer needed. In particular, transient soft robots have unique applications as unrecoverable and vanishing robots for hardware security in military operations like scouting, invasion, or transport, without being exposed to enemies (*13,14*); exploration of dangerous location such as deep sea or radioactive areas where the retrieval of robots is expensive or unavailable (*15-19*). However, the biggest bottleneck in the disposal of soft robots is the management of their frame material; a silicone elastomer. Although the low modulus, high stretchability, and processability of silicone elastomers make them a gold-standard frame material of soft robotics, the disposal of silicone elastomer is difficult owing to their highly crosslinked network structure, which allows silicone elastomers to bear ~300°C of heat, acids or alkaline with extreme pH conditions, and other hazardous chemicals (*20,21*). This stability makes bear silicone elastomers unsuitable for decomposable robots, resulting in alternative application fields. In addition, degradable elastomers spanning natural and synthetic polymers have recently gained high interest.

Previous studies on fully degradable soft robotic materials depended on passive transient elastomers such as poly(glycol sebacate) (PGS) and gelatin-based biogels. The PGS-CaCo$_3$ polymer was utilized to fabricate accordion-style pneumatic actuators and demonstrated full degradability after being buried in compost. (*22*) Gelatin biogel has also been applied to e-skins, soft actuators, and electronic sensor patches with excellent tunability, mechanical properties, and transience in deionized water. (*23*) However, these examples are still considered passive degradation considering the degradation is based on the hydrolysis rate of the materials. This type of robotic material degradation possesses two major drawbacks: 1) the performance of the materials or systems continuously degrades along with their decomposition and 2) the system lifetime is predetermined by the initial thickness and the environmental condition, usually without further controllability (*24*).

In contrast, on-demand transient soft robotic materials, which are highly stable yet rapidly degradable under specific external stimuli, are expected to overcome such drawbacks with their innate stimulus-responsive degradability. Unlike passive transient robots that undergo degradation immediately upon operation owing to immediate interaction with the surrounding environment, on-demand transient robots can initiate the degradation at a certain time frame without functional deterioration during operation, thereby enabling novel tasks such as 1) high and stable performance of robots without functional degradation during their operation period, 2) self-directed disappearance of temporary biomedical actuators after treatment, and 3) the programmable lifecycle of special-purpose robots. Self-immolative polymers, whose chains disintegrate under external conditions owing to their weak chemical bonding energy (*25*), are considered potential candidates for on-demand transient robotics.

However, such materials possess poor flexibility and stretchability and are unstable under ambient conditions owing to their low ceiling temperature ($T_c$) (*26,27*). For example, polyphthalaldehydes (PPAs) are the most widely used self-immolative polymers but possess a Young's modulus of 2.5–4.5 GPa, fracture strain of 1.5%–2%, and low ceiling temperature ($T_c$) of -45°C (*28*).

Herein, we propose on-demand transient and hyper-elastic robotic materials for life-time configurable soft robots by dispensing a photo-induced fluoride-generating diphenyliodonium hexafluorophosphate (DPI-HFP) to various silicone resins (Sylgard-184 and Ecoflex 00-30). In addition to the hyper-elasticity and simple processability that original silicone resins possess, the DPI-HFP/silicone composite realizes ultraviolet (UV) triggerable transient platform for soft robotics. Spectroscopic and thermal analyses revealed the degradation mechanism and kinetics of the Si–O–Si backbone cleavage initiated by fluoride ion ($F^-$) generated due to exposure to UV light. Furthermore, we fabricated a highly deformable and fully degradable gaiting robot and demonstrated it in a hypothetical scouting scenario. We integrated various electronics devices such as strain, temperature, and UV sensors to measure the external signals. When the robot was exposed to UV light (365 nm), it completely disintegrated, leaving behind an oily liquid residue of decomposed silicone composite and thin film electronics; as a result, it can no longer operate. The material presented in this study demonstrates the effective translation of conventional Si-O based silicone soft robotics to a life-cycle controllable form by combining them with an on-demand transient DPI-HFP/silicone composite.

**RESULTS**

**Overview of transient elastomer composites and lifetime-configurable soft robots**

Soft robots constructed with transient silicone composites can not only achieve diverse actuations owing to the superior mechanical properties but also have a configurable lifetime based on the disintegration of the material owing to certain external stimuli. Therefore, we established a strategy for designing transient silicone composites, including fabrication, application, and disintegration (Fig. 1). The on-demand transient material was fabricated by adding fluoride ion-emitting diphenyliodonium hexafluorophosphate (DPI-HFP) to uncured commercial silicone resin (Ecoflex 00-30 and Sylgard-184) (Fig. 1A). The desired morphology of the DPI-HFP/silicone composite was achieved by pouring the DPI-HFP/silicone mixture into a polylactic acid (PLA) mold and curing at 60°C for 30 min. Upon exposure to UV light, the DPI-HFP releases $F^-$ ions, cleaves the Si-O backbone, and converts the composite into an oily liquid. A gating robot composed of DPI-HFP/silicone was constructed as a representative application scheme (Fig 1B). Under certain scenarios wherein disintegration is desired, including mission completion, discovery by enemies, or disposal requiring volume reduction, the robot can be exposed to UV light and disintegrate into an unrecoverable form.

The sequential photographs shown in Fig. 1C demonstrate the on-demand transient behavior of the DPI-HFP/silicone gaiting robot. Applying UV light (365 nm, 30 min) followed by heat (120°C, 60 min) leads to complete decomposition through a phase transition from a crosslinked solid to a de-crosslinked oily liquid state. External heat is first transferred from the underlying hot plate to the robot feet and then dissipated throughout the body of the robot. The last image shows the complete degradation of the on-demand transient DPI-silicone gaiting robot, decomposing into an oily liquid state.

**Decomposition chemistry of DPI-HFP/silicone composites**

Fig. 2 shows the decomposition mechanism of DPI-HFP/silicone composite stated in Fig. 1 via spectroscopic analysis. Fig. 2A illustrates the decomposition mechanism, wherein the key to on-demand transience lies in the formation of fluoride ions from UV-exposed DPI-HFP. During decomposition, the diphenyliodonium cations of DPI-HFP undergo homolytic cleavage upon UV irradiation and yield phenyliodonium radical cations (*29*). These phenyliodonium radical cations attack and receive the protons available in their environment to form iodobenzene cations (*30*). The $PF_6$ anions of DPI-HFP remove the protons from iodobenzene cations, leading to the formation of $HPF_6$ (*31,32*). Because $HPF_6$ is highly unstable under ambient conditions, it naturally decomposes and yields protons, fluoride ions, and $PF_5$ as the final products of photolysis (*33,34*). Here, the concentration of fluoride ions generated from DPI-HFP at a given UV exposure time is predictable in principle (details are provided in Supplementary Note 1). These fluoride ions cleave the Si–O backbones of silicone elastomers as a result of a thermodynamic driving force; in other words, the bond energy of Si–F (565 kJ/mol) is greater than that of Si–O (452 kJ/mol) (*35*). This leads to the fragmentation of the organosilicon network of the silicone elastomer matrix and converting the solid elastomer into an oily liquid.

Figs. 2B-G (DPI-HFP/Ecoflex 00-30) and Supplementary Figs. 1A-C (DPI-HFP/Sylgard-184) show the spectroscopic observations of DPI-HFP/silicone composite before and after the trigger. The T2 relaxation time of DPI-HFP/Ecoflex 00-30 obtained using solid-state $^1H$ NMR relaxometry was used to compare the crosslinking density of the silicone elastomer before and after UV exposure (Fig. 2B). An increase in the $T_2$ relaxation time of a polymer, which is inversely proportional to the crosslinking density, indicates that the UV light leads to the decrosslinking and fragmentation of the organosilicon network (*36*). The FT-IR spectra in Fig. 2C show a decrease in the Si–O–Si bond peak (1013 $cm^{-1}$) after UV light

exposure owing to the Si-O bond cleavage via fluoride ions. In addition, the Si–CH$_3$ bond peak (792 cm$^{-1}$) decreased owing to the Pt catalyst included in the silicone elastomers for addition curing (*37,38*). The $^{29}$Si NMR spectra in Figs. 2D and 2E show a decrease in the in-chain Si peak (-22.56 ppm) and end-of-chain Si peak (6.56 ppm), respectively, owing to the Si-O bond cleavage. Comparing the $^{29}$Si NMR spectra in Figs. 2F and 2G showed changes in the chemical bonds before and after the UV trigger. Only Si-O bond peaks (-23 ppm) were observed in the $^{29}$Si NMR spectra before the UV trigger (Fig. 2F). As UV exposure generated fluoride ions, which attack the Si-O bonds, Si-F bond peaks (-8 ppm) and Si-OH bond peaks (-11.6 ppm) emerge in addition to the Si-O bond peaks (Fig. 2G). Additional $^{29}$Si -NMR analyses conducted on Sylgard-184 (fig. S2B) indicated similar trends before and after the UV trigger.

Considering the components and compositions of Ecoflex 00-30 and Sylgard-184 are proprietary and include significant yet uncertain amounts of additive materials (e.g., silica) that make the residue insoluble in organic solvents (*39*), this study used a model system of DPI-HFP/PDMS in analogy to DPI-HFP/Ecoflex 00-30 and DPI-HPF/Sylgard-184. $^{29}$Si NMR analyses on linear PDMS (Mn = 139,000) exhibited a similar trend before and after the UV trigger (fig. S2A). Additional analyses of the residue resulting from the model system comprising pure PDMS and DPI-HFP revealed the exact components generated by the trigger (figs. S3A-B). The FT-IR spectrum in fig. S3A shows that DPI-HFP/PDMS underwent the same Si–O–Si and Si–CH$_3$ cleavage as DPI-HFP/Ecoflex 00-30 (Fig. 2C) and DPI-HPF/Sylgard-184 (fig. S1A). The $^{29}$Si NMR spectra in fig. S3B indicates the exact components and ratios resulting from the trigger: cyclic siloxane compounds octamethylcyclotetrasiloxane (D4), decamethylcyclopentasiloxane (D5), and dodecamethylcyclopentasiloxane (D6) and shortened hydroxyl- and fluorine-terminated PDMS. Furthermore, the formation of shortened linear PDMS chains and cyclic siloxane compounds was confirmed from the decrease in the average molecular weight from 10$^4$–10$^5$ g/mol to 10$^2$–10$^3$ g/mol, as measured by gel permeation chromatography (GPC) performed before and after the application of the trigger (fig. S3B). The formation of hydroxyl- and fluorine-terminated PDMS is consistent with the mechanism suggested in Fig. 2A, and their further backbiting-mediated decomposition into cyclic siloxane compounds is consistent with previous results on the decomposition of PDMS with fluoride species (*39,40*). The same compounds appeared in the residues of DPI-HFP/Ecoflex 00-30 (Fig. 2G) and DPI-HFP/Sylgard-184 (fig. S2B), although the cyclic siloxane compounds were not visible owing to the innate peak broadening and resolution limit of solid-state NMR. In contrast, cyclic siloxane compounds are recyclable (*39,41*), and shortened, fragmented PDMS chains are considered suitable for soil or clay mineral-induced degradation (*42,43*). Overall, the spectroscopic observations of DPI-HFP/Ecoflex 00-30 (Fig. 2B–G) and DPI-HFP/Sylgard-184 (figs. S1A–C) indicate that the material underwent Si–O cleavage by the fluoride ion generated from DPI-HFP.

**Decomposition kinetics of DPI-HFP/silicone composites**

Verification of the kinetic parameters related to decomposition helps predict the degradation profile of the DPI-HFP/silicone composites, thereby enabling the design of transient silicone composites with diverse lifetimes. Therefore, kinetics analysis was performed under various thermal conditions to characterize the decomposition behavior of the DPI-HFP/silicone composite. Fig. 3A and fig. S4 depict the decomposition of the DPI-HFP/Ecoflex 00-30 and DPI-HFP/Sylgard-184, respectively. Decomposition was triggered by applying 365 nm UV light for 30 min. Then, the decomposition rate was accelerated by heating at 120°C. In the decomposition process, the composites undergo a phase change from a crosslinked solid to a de-crosslinked oily liquid. Thermal analysis of the DPI-HFP/Ecoflex 00-30 composite using photo-differential scanning calorimetry (photo-DSC) was performed to determine the enthalpy required to decompose the composite and calculate the decomposition rate acceleration under various conditions (Figs. 3B-D and figs. S5A-E). Without applying the UV trigger, no heat flow was observed even at an elevated temperature of 120°C (Fig. 3B). In contrast, when UV light was applied (Fig. 3C), heat flow (~15,000 s) was observed even at room temperature (25°C) owing to the occurrence of an exothermic Si-O cleavage reaction. When UV light and 120°C heat were both applied (Fig. 3D), accelerated heat flow was observed (~4500 s).

The extent of phase conversion ($\alpha$) of DPI-HFP/silicone can be measured via thermal analysis, considering the decomposition of DPI-HFP/silicone involves exothermic Si–O cleavage (*44*). Here, the value of $\alpha$ at time t is empirically quantified as:

$$\alpha = \frac{\Delta H_t}{\Delta H_{total}}, \quad (1)$$

where $\Delta H_{total}$ is the total amount of heat emitted during the decomposition and $\Delta H_t$ is the amount of heat emitted until time t. Photo-DSC analysis applied to Eq. (1) at various temperatures yielded decomposition profiles of DPI-HFP/Ecoflex 00-30, as shown in Figs. 3E and 3F. The decomposition of DPI-HFP/Ecoflex 00-30 was triggered by UV irradiation; therefore, external heating to 120°C without UV irradiation did not cause phase conversion. With prior application of a UV stimulus, the initial slope of the decomposition profile increased as the heating temperature increased, which indicated that external heat accelerates the decomposition reaction. This demonstrates that after photolysis, the overall reactions shown in Fig. 2A were accelerated owing to external heating because reactants with sufficient energy to overcome the activation energy barrier of the Si–O cleavage reaction became available.

Moreover, Vyazovkin's model (*45*) provides a theoretical expression relating $\alpha$ to the rate constant k:

$$\alpha = 1 - e^{-kt} \qquad (2)$$

Detailed derivations are provided in Supplementary Note 2, with Supplementary Fig. 5. Measuring α using photo-DSC and using it in Eq. (2) provides the value of k at each temperature (fig. S5), allowing us to generate an Arrhenius plot (Fig. 3G). The Arrhenius plot of the conversion rate, which is representative of the decomposition rate in the activation energy-based process, yields an Arrhenius pre-exponential factor A = 0.1703 and activation energy $E_a$ = 18.09 kJ/mol, which facilitate the prediction of the degradation profile of DPI-HFP/Ecoflex 00-30 under UV radiation at arbitrary temperatures. The phase conversion datasets and Arrhenius plot for DPI-HFP/Sylgard-184 are shown in Supplementary Fig. 6. Kinetic analysis of DPI-HFP/Ecoflex 00-30 (Fig. 3) and DPI-HFP/Sylgard-184 (fig. S6) show that these materials exhibit similar decomposition behavior.

The decomposition behavior of identical DPI-HFP/Ecoflex 00-30 composite cubes (0.8 × 0.8 × 0.8 cm$^3$) under normal sunlight and UV-triggered conditions without additional heat exposure was analyzed (fig. S7). The energy transferred from sunlight was not sufficient to overcome the activation energy barrier to trigger decomposition; therefore, the DPI-HFP/Ecoflex 00-30 did not undergo a phase change without the designated trigger of UV light, which plays a key role in designing of reliable soft robots in real-world applications.

**Application to a soft robotic actuator**

Hyper-elastic silicone elastomers have been utilized extensively in the design of soft robots, considering these materials allow various unconstrained movements. However, the stress concentration at the interface of silicone and DPI-HFP results in the formation of voids inside the material, which act as crack initiators and decrease the fracture strain (*46,47*). Incorporating a low volume fraction of DPI-HFP (maximum 11.8 vol%, 20 wt% DPI-

HFP/silicone elastomer) and a material preparation process using finely powdered DPI-HFP allows abundant network formation of the filler with the silicone elastomer matrix to retain the elastic modulus (*48*). We verified the conservation of the hyper-elastic properties of the transient DPI-HFP/silicone composites via physical distortion and biaxial elongation tests, and fabricated a gaiting soft robot. The physical distortion tests showed that the DPI-HFP/silicone possessed hyper-elastic and tear-resistive mechanical properties (Fig. 4A) that originated from the silicone matrix. The tensile stress–strain evaluation of Ecoflex 00-30 and Sylgard-184 (Fig. 4B and figs. S8 and S9) showed that the composites undergo large elastic deformations before failure (Fig. 4B, DPI-HFP/Ecoflex 00-30; fig. S8, DPI-HFP/Sylgard-184). The addition of DPI-HFP to Ecoflex 00-30 decreased the fracture stress and strain (0 wt%, 0.4251 MPa and 683.72%; 10 wt%, 0.1453 MPa and 571.67%; 20 wt%, 0.1897 MPa and 493.34%) but did not change the elastic modulus (400% elastic limit, 40.02 ± 1.65 kPa elastic modulus). Such mechanical properties are suitable for soft robotic applications and are highly superior to those of other stimulus-responsive composite materials reported, such as cPPA/PAG (*28*).

We designed and fabricated a destructible-on-demand pneumatic gaiting soft robot that takes advantage of the hyper-elastic properties of the DPI-HFP/silicone composites. Fig. 4C shows a schematic of our assembled robot (left) and an exploded view of its actuation components (right). The robot comprised a bending actuator with an air channel (DPI-HFP/Ecoflex 00-30), a strain limiter (DPI-HFP/Ecoflex 00-30), and four supporting feet (DPI-HFP/Sylgard-184). The dimensions of each component are shown in fig. S10. Fig. 4D shows photographs (top) and corresponding finite element analysis (FEA) strain profiles (bottom) of the robot during pneumatic-actuated motion under pressures between 12 kPa to -12 kPa. Applying pneumatic pressure to the air channel caused a bending motion owing to the difference in thickness, and in turn the stiffness between the bending actuator and the strain limiter. The relatively rigid feet supported the structure robustly during movement and aided the robot in moving forward (*49*). When the robot was in a flexed state, the rear feet showed a higher frictional force compared to the front feet owing to its larger area of contact with the ground. As a result, the robot moved forward as it transitioned into an extended state owing to greater displacement at the front feet.

FEA provided quantitative values for the actuation mechanism (Fig. 4D, cut view; fig. S11, bare view). The distribution of principal strains upon pressurization showed that a maximum strain of 83.56% at the top wall of the air channel when the robot was in a flexed state. The bending angles of the pressurized Ecoflex 00-30 and DPI-HFP/Ecoflex 00-30 actuators were measured experimentally to demonstrate that adding a DPI-HFP does not alter the original performance of a pure Ecoflex 00-30 (Fig. 4E). The Ecoflex 00-30 and DPI-HFP/Ecoflex 00-30 actuators exhibited similar bending angles at the same pressure. We observed the displacement of a DPI-HFP/silicone gaiting robot under cyclic pressure between 12 kPa to -12 kPa (Fig. 4F). As a result, the robot moved steadily at a velocity of 2.5 cm/s through repeated extension and flexion.

**Direct integration of multipurpose soft electronics**

The integration of a multipurpose soft electronic system with a pneumatic soft robot expands the on-demand transient robot functionality to ambient condition monitoring, destructive condition detection and self-alarm, and actuation control. A temperature sensor array, strain sensor, and photodetector array were integrated with an on-demand transient robot fabricated from DPI-HFP/silicone composites. The strain sensor collected information about the strain induced during the walking motion of the robot and provided feedback about the motion data to control the robotic movement. The temperature sensor detected the ambient temperature, and the photodetectors detected ultraviolet (UV) light during regular operating conditions. Both the temperature sensor and photodetector served to monitor for the triggering conditions that initiate the destruction of the robot.

Fig. 5A shows an image of the actual strain sensor and temperature sensor array integrated into the strain limiter of the robot and exploded view schematics. Both sensors comprised ~ 300 nm thick copper layer deposited on a supportive polyimide (PI) dielectric film (~ 10 µm thick). Fig. 5B shows the actual UV sensor array comprising PIN photodiodes installed on the forehead of the robot and its exploded view schematic; the array comprised a monocrystalline silicon membrane (~ 1500 nm thick) with a channel length of 20 µm and width of 625 µm and Cu (~ 300 nm thick) electrodes.

Fig. 5C shows the fabricated strain sensor, which determines the strain by measuring the capacitance of the interdigitated electrodes, and a magnified view. Fig. 5D shows the capacitance measured during a single cycle of robot walking, where the bending angle ranged from 0–35°. The sensors were designed such that an increase in capacitance was measured as the distance between the two neighboring electrodes increased (*50*). The strain induced in the region where the strain sensor was located was analyzed via FEM analysis, and the experimental data of the capacitance change was compared with the FEM data of the strain change at different bending angles (fig. S13A). fig. 13B shows the capacitance converted to strain. Fig. 5E shows the continuous capacitance measurements of the strain sensor during robot walking. The capacitance change was consistently 1 pF under cyclical pneumatic pressurization between -10 kPa and 10 kPa.

Fig. 5F shows a magnified view of the temperature sensor array using the thermal resistivity of the material (*51*). The measured temperature coefficient of resistance (TCR) of the copper was 2 mΩ·°C$^{-1}$ in the temperature range of 25–100°C (Fig. 5G), which is comparable to the value reported in the literature (3.69 mΩ·°C$^{-1}$) (*52*). Joule-heated rods at temperatures of 50, 70, and 100°C were placed under sensors 1, 3, and 5, respectively, and the temperatures across the surface resulting from heat diffusion were measured using an infrared camera (Fig. 5H, left) and the temperature sensor array (Fig. 5H, right). Temperature variations throughout the robot body, measured via the integrated temperature sensor array, corresponded to the calibrated temperature measurement shown in Fig. 5H (*53*).

Fig. 5I shows the Si PIN photodiode array and a magnified photodiode, which utilize the photoelectric current of a p-n junction to detect the UV light (*54*) that triggers the robot to disintegrate. Fig. 5J shows the current-voltage (I-V) characteristics in both dark (without UV light) and bright (365 nm UV light) modes. The currents induced in dark and bright modes were 0 A and $3.4 \times 10^{-8}$ A, respectively, under the applied voltage sweep from -2 V to 2 V. When the UV light was cycled through on/off modes (0 V to -2V), the induced photocurrent was consistently $-5 \times 10^{-8}$ A at -2 V (Fig. 5K).

**Demonstration of a lifetime configurable gaiting robot**

We operated the gaiting soft robot with various electronics under an arbitrary set of conditions and ultimately triggered its disintegration. To demonstrate its functionality, the robot was deployed on a series of hypothetical military missions, including scouting a foreign environment without exposure to unwanted parties, recognizing the decomposition risk factors, escaping to avoid destruction, and finding a triggerable environment for the transient DPI-HFP/silicone-based gaiting robot to self-disintegrate (Figs. 6B and C, Movie S1). First, the robot entered Zone 1 (60°C via a heat gun) and collected information regarding the temperature of the environment. Then, the robot entered Zone 2 (365 nm UV light on) and collected information regarding the UV light in the environment. The material transience began at this point due to the UV light trigger. When the robot entered Zone 3 (120°C via hot plate), it noted the hazardous risk of accelerated decomposition owing to the high temperature and warned the operator to escape the zone quickly. After the entire mission was complete, the robot entered the final zone (120°C via a hot plate) and rapidly decomposed into an oily liquid state within 1 h, leaving no potential for recovery.

Fig. 6A shows the before and after images with respect to the destruction of the interfaced robot after the trigger (365 nm UV light, 120°C heating). The top row presents an integrated view of each sensor in the DPI-HFP/silicone robot. The bottom row shows images of the strain sensor, temperature sensors, and photodetectors that collapsed and entangled in an unusable and non-restorable form with the destruction of the robot body. Furthermore, the electronic failure was assisted by the dissolution of the Cu used for the metal electrodes induced by the production of F$^-$ ions during decomposition (*55*). The functional failures of the sensors to the nonrecoverable form, induced by the destruction of the robot, are illustrated in Supplementary Fig. 15. fig. 15A shows a robust change in the strain sensor capacitance before failure owing to robot body instability during decomposition. fig. 15B shows the stable resistance reading of 0 Ω from the temperature sensor before the robot was triggered, which rapidly exceeded 106 Ω as the robot was triggered with UV light and heat. Additionally, Supplementary fig. 15C shows that the induction of photocurrent (-5 × 10$^{-8}$ A at -2 V) was disabled as the robot collapsed.

**DISCUSSION**

The proposed DPI-HFP/silicone composites enable a novel mechanism for on-demand material decomposition initiated by UV exposure while maintaining the highly elastic and stretchable mechanical properties of conventional silicone elastomers used in soft robots. FT-IR and NMR analyses after UV exposure revealed the sequential chemistry of the photo-generation of F- ions from DPI-HFP and the decomposition of Si–O–Si bonds, which indicated the backbone cleavage mechanism of silicone composites owing to the generated fluoride ions. The temperature-dependent kinetics were analyzed by photo-DSC, and the increasing rate of accelerated decomposition with the increase in temperature was determined based on the Arrhenius equation.

A largely deformable and on-demand disintegrating pneumatic gaiting soft robot was designed and fabricated by combining DPI-HFP/Ecoflex 00-30 and DPI-HFP/Sylgard-184 composites, thereby demonstrating the usability of our novel material for soft robotics applications. Embedding the robot with flexible electronics in the form of strain, temperature, and UV light sensors successfully provided multifunctionality for controlling robotic motion, scouting the surrounding environment, and self-diagnosing a triggerable condition to choose self-protection or self-destruction.

The concept and materials illustrated in this study have extensive potential applications, such as in remote-controlled military applications, hardware-secured devices, robotic exploration of hazardous locations, and effective waste processing. In addition, further chemistry studies on these DPI-HFP/silicone composites can help reduce the trigger time scale or the repolymerization of the residual liquid and expand the field of on-demand transient robotics.

## MATERIALS AND METHODS

**Preparation of DPI-HFP/silicone composite**

The required amount of diphenyl iodonium hexafluorophosphate (DPI-HFP; TCI, Japan) for a 1:1 mass ratio was added to Ecoflex 00-30 prepolymer (Smooth-On, USA), followed by manual stirring using a metal stick to prepare DPI-HFP/Ecoflex 00-30 resin. Then, the mixture was placed in a vacuum desiccator for 5 min to remove air bubbles. The mixture was poured into a 3D-printed mold and cured in an oven at 60°C for 30 min. The same procedure was followed to prepare the DPI-HFP/Sylgard-184 composite using Sylgard-184 (Dow Corning, USA), except the prepolymer was in a 20:1 ratio and curing was performed for 60 min. To prepare the DPI-HFP/PDMS composite used for residue analysis, trimethyl-terminated PDMS (M.W. = 139,000; Alfa Aesar, USA) was mixed with 20 wt% DPI-HFP.

**Characterization and decomposition analysis of DPI-HFP/silicone composites**

To determine the mechanical properties of the DPI-HFP/silicone composites, uniaxial tensile tests were performed using an Instron 3343 universal testing machine (Instron, USA) with a fixed receding strain rate of 10%/s. Nicolet iS50 FT-IR spectrophotometer (Thermo-Fisher Scientific, USA) was used for decomposition chemistry analysis. $^{29}$Si and $^{1}$H solid-state NMR data were obtained using a 500 MHz Avance III system (Bruker, Germany), and solution-state $^{29}$Si-NMR data were obtained using a 600 MHz Avance 600 system (Bruker, Germany) with THF as the solvent. Gel permeation chromatography (GPC) measurements were obtained with a Shodex SEC LF-804 column with a Wyatt OptiLab T-rEx refractive index (RI) detector, with chloroform as the solvent. Decomposition kinetics were measured using a photo-DSC system comprising a DSC-Q200 (TA Instruments, USA) and an Omnicure-s2000 spot-cure light source (Excelitas, USA).

**Fabrication, characterization, and simulation of soft robots**

The robot bending actuator and feet were formed by curing DPI-HFP/Ecoflex 00-30 and DPI-HFP/Sylgard-184 in 3D printed PLA molds. Then, the uncured DPI-HFP/Ecoflex 00-30 was applied at the interfaces between each part of the integrated robot and cured at 60°C for 30 min to act as an adhesive. A small hole was drilled in the body using a needle, and a tube was inserted for pneumatic actuation. The displacement and bending angle of the robot were characterized using image analysis software (Tracker 5.1.5, Open Source Physics), whereas the pneumatic pressure was measured using a customized Arduino setup comprising a pressure sensor and a microcontroller.

The motions of the transformable robot were simulated via 3D finite element analysis (FEA) using commercial software (ABAQUS, Dassult Systemes, France). The main body part of the transformable robot was developed with an inner empty space to model air pressure. In the empty space, a uniform pressure was applied to the walls perpendicular to the inner surface without air injection holes. The element type was a 4-node linear tetrahedron (C3D4). The elastic modulus, Poisson's ratio, and density were set to 40 kPa, 0.43, and 1.07 g/cc, respectively, which corresponded to the mechanical properties of DPI-HFP/Ecoflex 00-30.

**Fabrication of multipurpose electronic sensors**

figs. S12 and S14 illustrate the electronics fabrication processes for the temperature, strain, and UV light sensors. The fabrication of temperature sensors and strain sensors began by laminating a polyimide (PI) film (10 µm, Goodfellow) onto a DPI-HFP/Ecoflex 00-30 (~500 µm)-coated glass slide. A thin film of copper (Cu, 300 nm) was deposited on the PI via sputtering. Laser cutting of the PI-Cu bilayer defined the resistive structure for the temperature sensor and capacitive structure for the strain sensor. The DPI-HFP/Ecoflex 00-30 with embedded sensors was removed from the glass slide and adhered onto the DPI-HFP/Ecoflex 00-30 robot.

To fabricate the photodiode array, a $SiO_2$ diffusion mask was deposited on an SOI wafer (top Si thickness ~1500 nm, p-type, Soitec) via plasma-enhanced chemical vapor deposition (PECVD). The diffusion doping of boron (spin-on-dopant 20B, Filmtronics; tube furnace at 1050°C with $N_2$ flow) and phosphorous (spin-on-dopant P509, Filmtronics; tube furnace at 1000°C with $N_2$ flow) through the $SiO_2$ diffusion mask yielded Si membrane PIN diodes. The buried oxide was removed by applying HF acid into the hole pattern (diameter ~5 µm) that was dry etched with reactive ion etching ($SF_6$, 15 sccm, 200 W). The removal of the buried oxides allowed the top monocrystalline Si membrane to be released and transferred onto a diluted PI (D-PI) membrane (converted from poly(pyromellitic dianhydride co-4,4-oxydianiline), ~1.5 µm, Sigma–Aldrich) on a sacrificial PMMA-coated (poly(methyl methacrylate), ~600 nm, Microchem) silicon wafer. Electrical connections and contact pads (Cu, ~300 nm) were formed by sputtering and lift-off procedures. D-PI was spin-coated over the diode array and photopatterned to dry etch the excessive D-PI layout. Furthermore, PMMA removal via immersion in acetone allowed the PIN diode array to be released and transferred onto the DPI-HFP/Ecoflex 00-30 robot body.

**Supplementary Materials**

Supplementary Text 1: Kinetics of the photolysis of DPI-HFP

Supplementary Text 2: Details of the Vyazovkin's model

Fig. S1. Chemical structure analysis of DPI-HFP/Sylgard-184.

Fig. S2. $^{29}Si$ NMR analyses of decomposition residues.

Fig. S3. Component analyses of DPI-HFP/PDMS decomposition residue.

Fig. S4. Photo-triggered decomposition of a silicone elastomer by a photo-fluoride generator.

Fig. S5. DSC analysis of DPI-HFP/Ecoflex 00-30 at various temperatures.

Fig. S6. Kinetic analysis of DPI-HFP/Sylgard-184.

Fig. S7. Decomposition of silicone elastomer composites by a photo-fluoride generator.

Fig. S8. Mechanical properties of DPI-HFP/Sylgard-184.

Fig. S9. Tensile testing of Ecoflex composites with various DPI-HFP concentrations.

Fig. S10. Dimension schematics of the pneumatic walking robot.

Fig. S11. FEM-analyzed strain distributions of the gaiting robot.

Fig. S12. Schematic overview of temperature and strain sensor fabrication.

Fig. S13. Conversion of capacitance to strain.

Fig. S14. Schematic overview of photodiode array fabrication.

Fig. S15. Transient behaviors of electronics integrated with the decomposing pneumatic soft robot.

Movie S1. Hypothetical military operation of the on-demand transient soft robot.

**Acknowledgments**

**Funding:**

Korea Medical Device Development Fund grant funded by the Korea government (the Ministry of Science and ICT, the Ministry of Trade, Industry and Evergy, the Ministry of Health & Welfare, Republic of Korea, the Ministry of Food and Drug Safety) (Project Number: KMDF_PR 20200901_0205-02)

National Research Foundation of Korea (NRF) grant funded by the Korea government (MSIT) (NRF-2019R1C1C1004232)

KIST Institutional Program (Project Number: 2V07080-19-P141)

**Author contributions:**

Conceptualization: M.-H. Oh, Y.-H. Kim, S.-M. Lee, S. Y. Kim, S.-K. Kang

Methodology: M.-H. Oh, Y.-H. Kim, S.-M. Lee

Investigation: M.-H. Oh, Y.-H. Kim, S.-M. Lee

Visualization: M.-H. Oh, Y.-H. Kim, S.-M. Lee, G.-S. Hwang, K.-S. Kim, J.-Y. Bae, J.-Y. Lee, J.-Y. Kim, Y.-C. Kim

Supervision: S. Y. Kim, S.-K. Kang

Writing draft: M.-H. Oh, Y.-H. Kim, S.-M. Lee, G.-S. Hwang

**Competing interests:** Authors declare that they have no competing interests.

**Data and materials availability:** All data are available in the main text or the supplementary materials.

**Figures:**

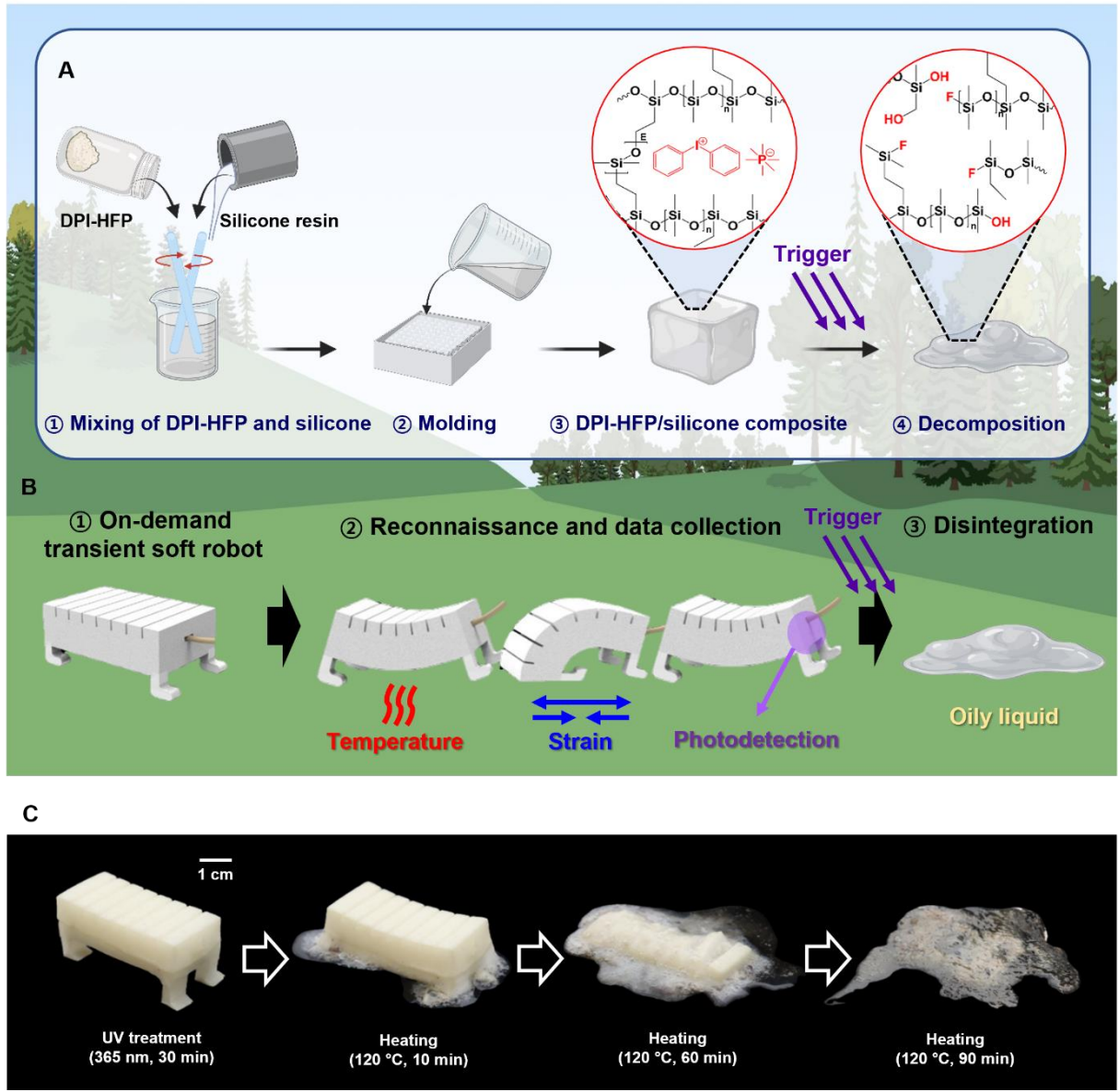

**Fig. 1. Overview of transient DPI-HFP/silicone composites and a -lifetime configurable soft robot.** (**A**) Fabrication process of a transient DPI-HFP/silicone elastomer composite and its decomposition response upon application of a trigger. (**B**) Illustration of a lifetime configurable gaiting robot capable of sensing the surrounding environment and disintegrating the entire system via trigger application whenever necessary. (**C**) Time-lapse image of the gaiting robot undergoing decomposition at 120°C for 30 min after exposure to UV light (365 nm).

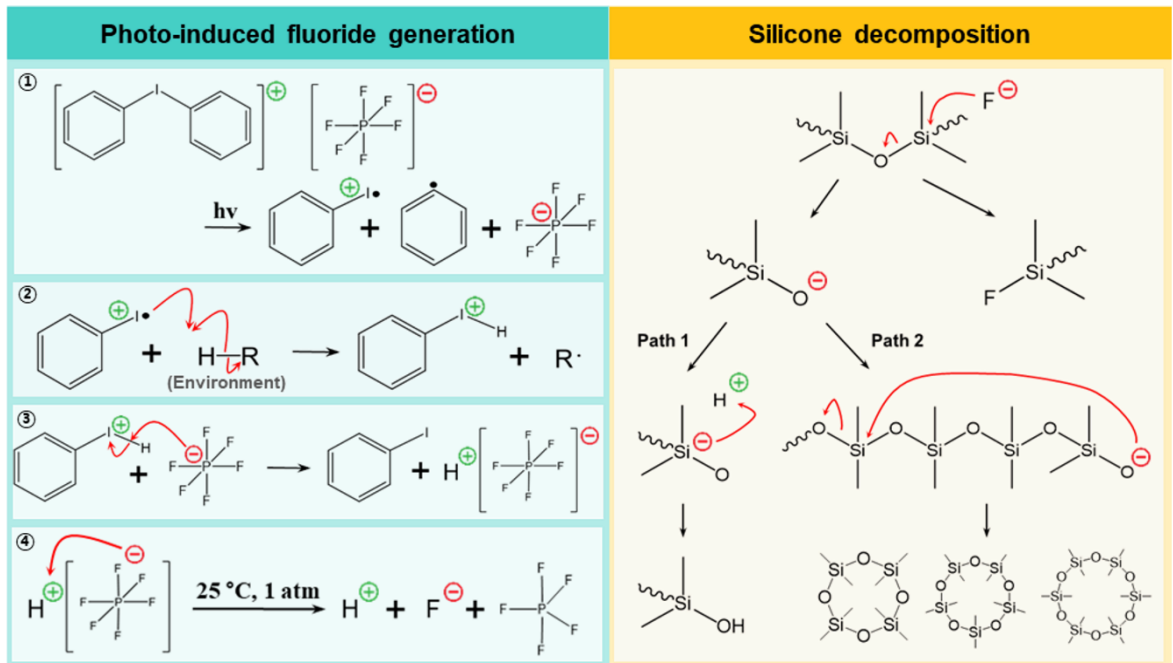
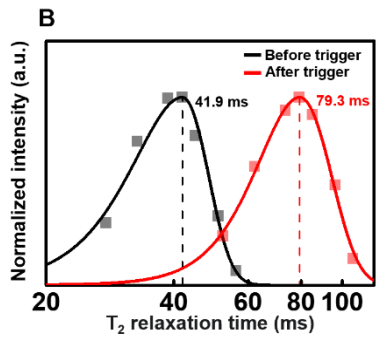
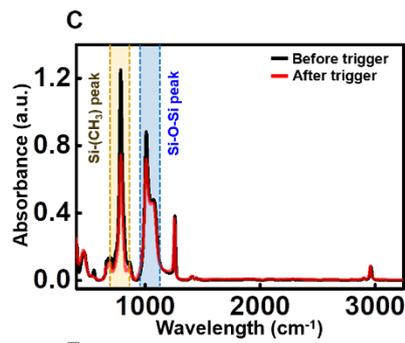
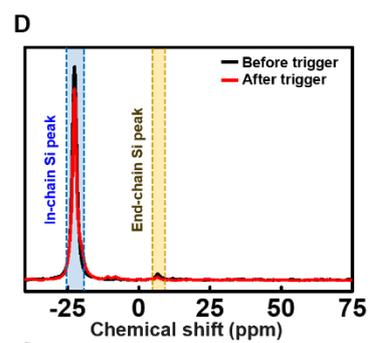
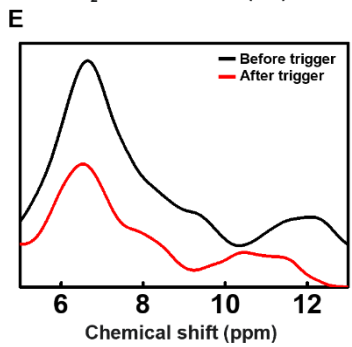
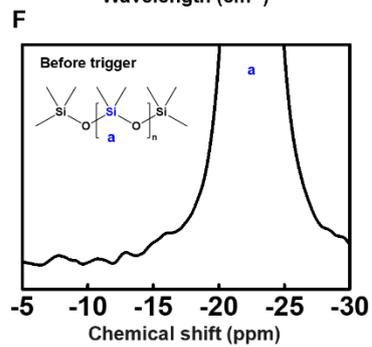
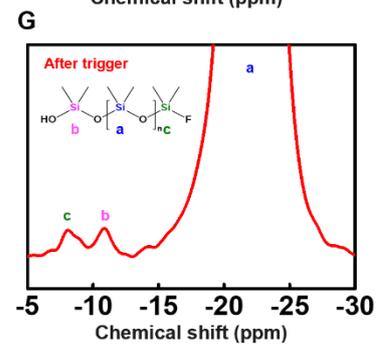

**Fig. 2. Decomposition chemistry of DPI-HFP/silicone composites that utilizes photoinduced fluoride.** (**A**) Schematic of the phototriggered decomposition of the DPI-HFP/silicone composites. The exposure of DPI-HFP to UV generates fluoride species (left), which subsequently results in Si–O bond cleavage to decompose the silicone elastomer (right). (**B**) Transverse relaxation time ($T_2$) distributions of DPI-HFP/Ecoflex 00-30 using solid-state $^1$H-NMR. The $T_2$ relaxation time increases after applying UV and heat trigger. (**C**) FT-IR spectra of DPI-HFP/Ecoflex 00-30 shows an absorbance decrease in the Si–CH3 peak (792 cm$^{-1}$) and the Si–O–Si peak (1013 cm$^{-1}$) after applying the trigger. (**D**) Solid-state $^{29}$Si NMR spectra of DPI-HFP/Ecoflex 00-30 showing that the intensity decreases in the Si–CH$_3$ peak (-22.56 ppm) and Si–O–Si peak (6.56 ppm) after applying the trigger. (**E**) Magnified and rescaled view of the $^{29}$Si-NMR spectra of DPI-HFP/Ecoflex 00-30 after applying the trigger. A significant decrease is visible in the Si-O-Si peak (6.56 ppm). (**F**) Solid-state $^{29}$Si-NMR spectra of DPI-HFP/Ecoflex 00-30 before application of the trigger. (**G**) Solid-state $^{29}$Si-NMR spectra of DPI-HFP/Ecoflex 00-30 after application of the trigger, showing the formation of cyclic hydroxyl- and fluoride-terminated PDMS chains. Here, peaks attributed to cyclic siloxane compounds are buried in the large Si-O-Si peak, and cyclic siloxane peaks are not observed owing to the innate peak broadening of solid-state NMR.

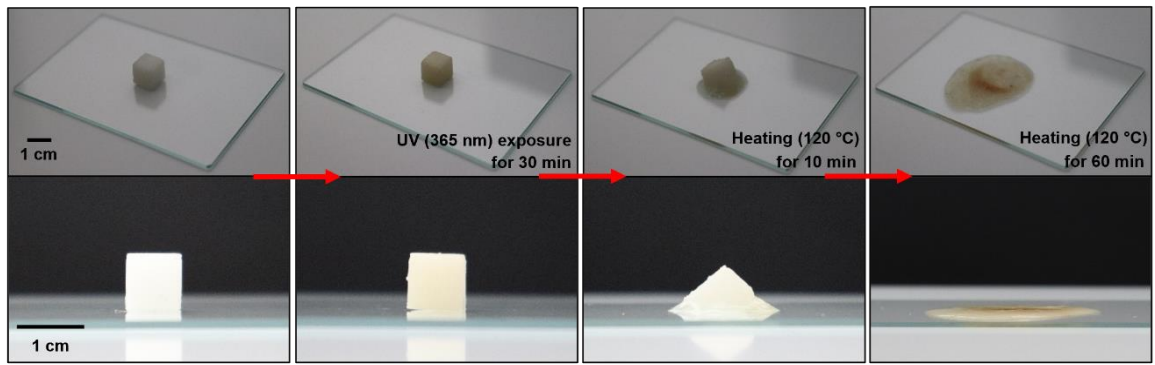
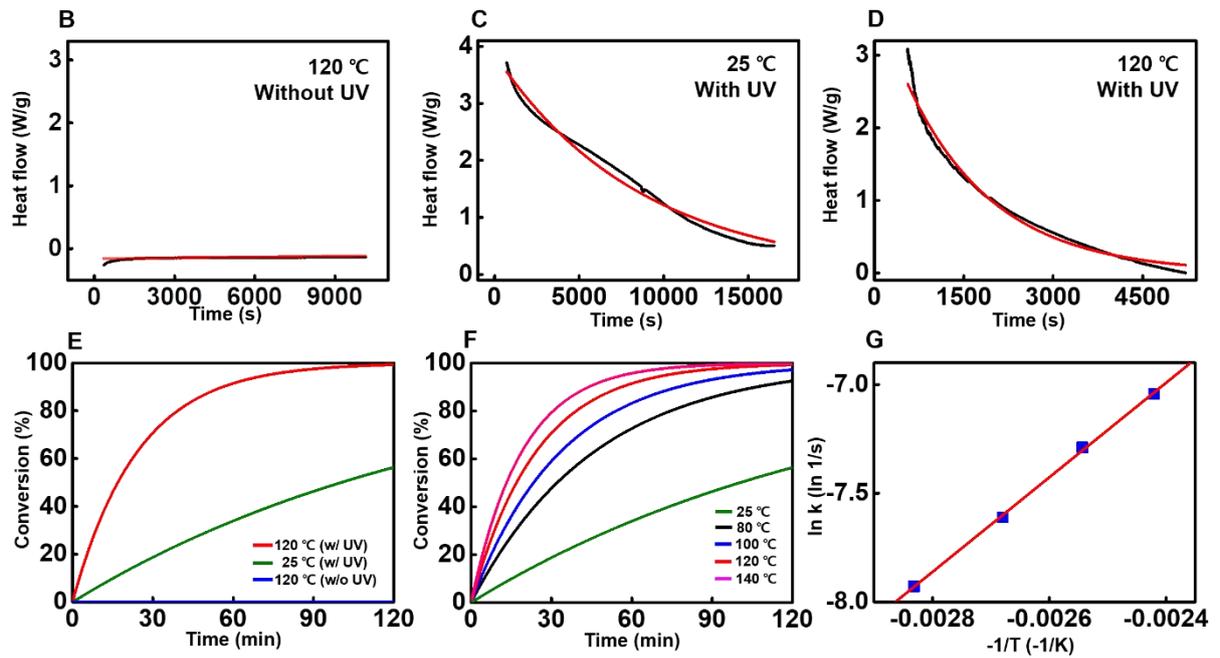

**Fig. 3. Decomposition kinetics of the DPI-HFP/silicone composite.** (**A**) Time-lapse photos showing decomposition of DPI-HFP/Ecoflex 00-30 triggered by exposure to UV (365 nm) for 30 min and followed by heating at 120°C for 1 h. (**B**) Photo-DSC curve of DPI-HFP/Ecoflex 00-30 at 120°C without UV exposure, revealing the stability of the composite without exposure to UV light (black, experimental data; and red, fitted data). (**C**) Photo-DSC curve of DPI-HFP/Ecoflex 00-30 at 25°C with UV exposure, revealing the UV-triggered transience of the composite (black, experimental data; red, fitted data). (**D**) Photo-DSC curve of DPI-HFP/Ecoflex 00-30 at 120°C with UV exposure, revealing the acceleration of decomposition upon exposure to heat (black, experimental data; red, fitted data). (**E**) DPI-HFP/Ecoflex 00-30 phase conversion under various conditions. Decomposition did not occur without exposure to UV light and increases as the temperature increases (120°C with UV, red; 25°C with UV, green; 120°C without UV, blue). (**F**) Phase conversion of DPI-HFP/Ecoflex 00-30 with UV radiation at different temperatures. Higher temperatures at decomposition resulted in faster decomposition rates (25°C, green; 80°C, black; 100°C, blue; 120°C, red; 140°C, magenta). (**G**) Arrhenius plot of DPI-HFP/Ecoflex 00-30 for calculating the kinetic parameters such as the activation energy (Ea) and reaction rate constant (k).

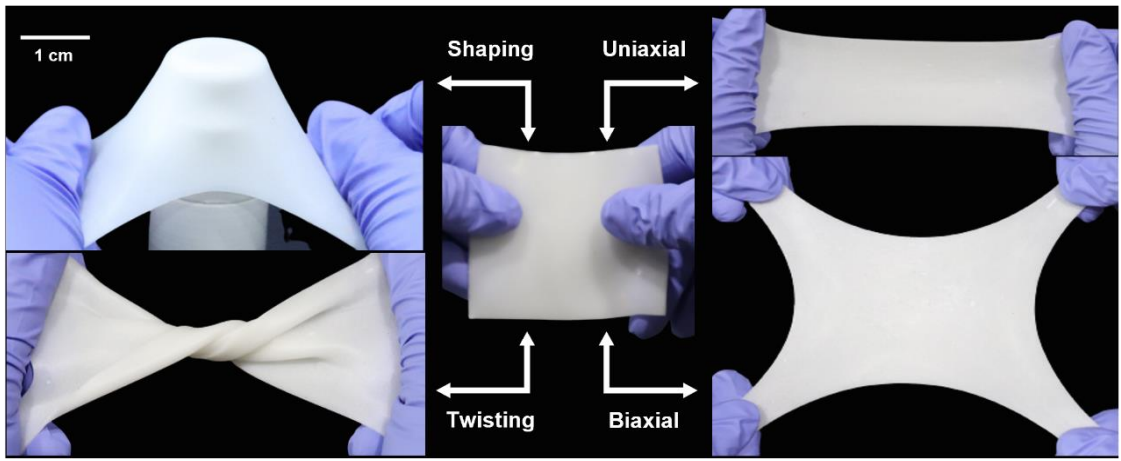

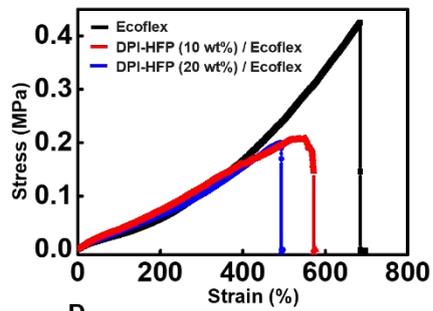
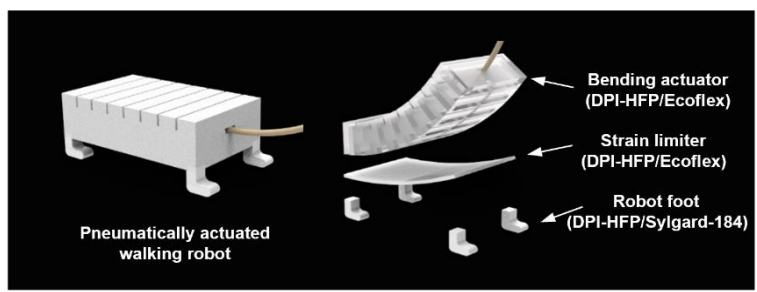

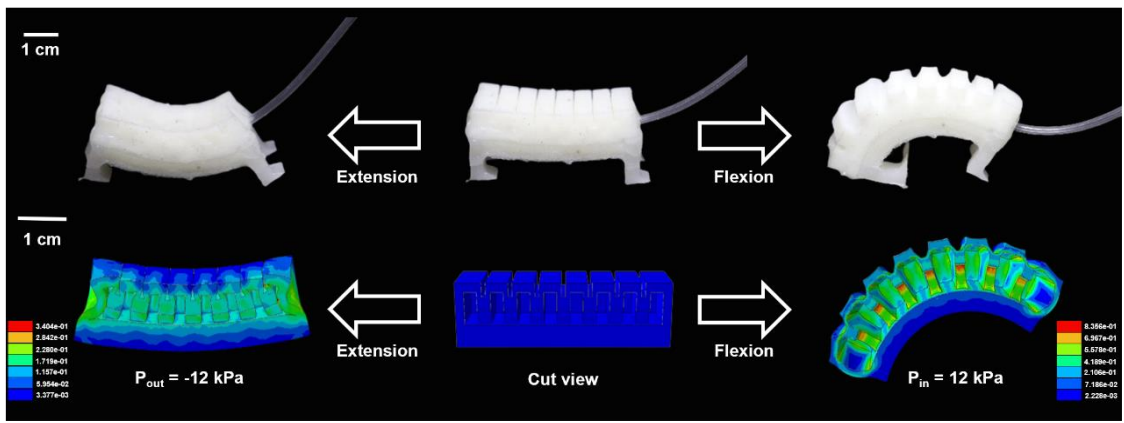

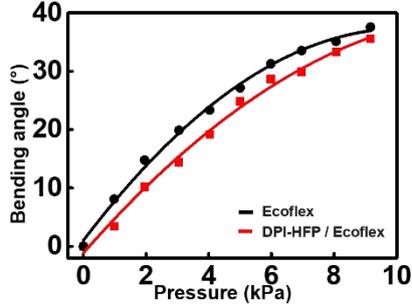
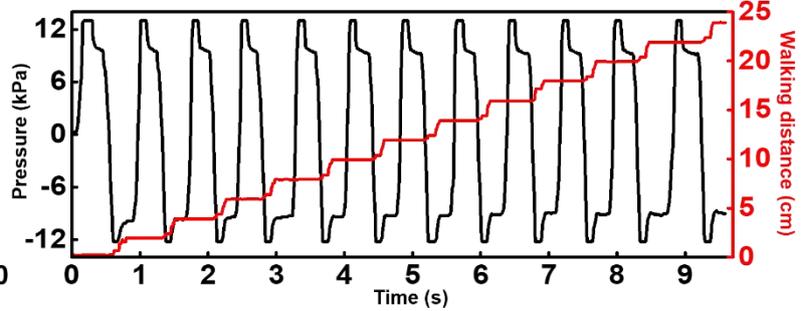

**Fig. 4. Hyper-elastic properties of DPI-HFP/silicone composites and application to a gaiting soft actuator.** (**A**) Photos demonstrating the hyper-elastic and tear-resistive properties of the composite under various strain conditions. (**B**) Stress–strain behavior of DPI-HFP/Ecoflex 00-30 composites at various DPI-HFP concentrations (0 wt%, black; 10 wt%, red; 20 wt%, blue). (**C**) Schematics of the assembled (left) and exploded view (right) of a pneumatic gaiting actuator. (**D**) Photographs and simulated FEM-analyzed internal strain distributions of the gaiting soft robot during extension (left) and flexion (right) through pneumatic actuation. (**E**)Experimental measurements of the bending angle of an Ecoflex 00-30 actuator as a function of the inlet pressure with (20 wt%, red) and without (black) added DPI-HFP. (**F**) Locomotion distance (red) of the gaiting robot as a result of the cyclic pneumatic pressure (black).

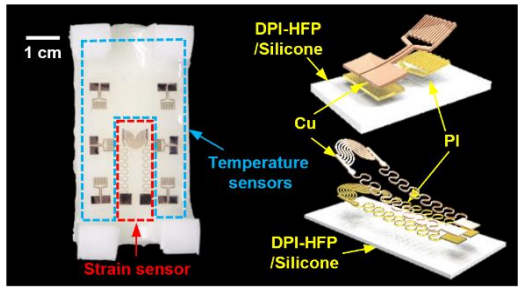
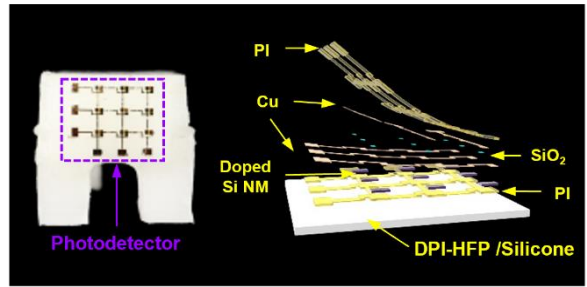
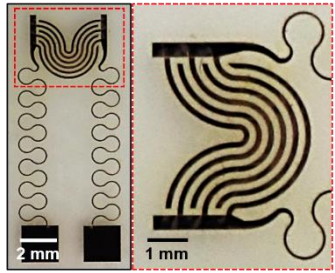
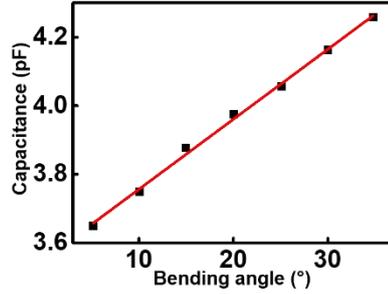
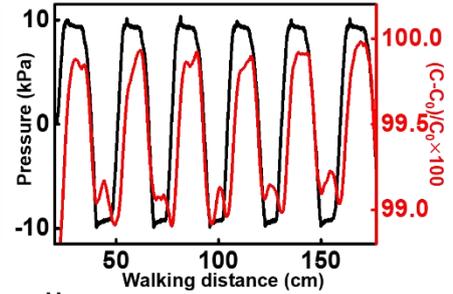
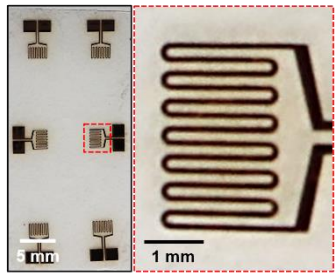
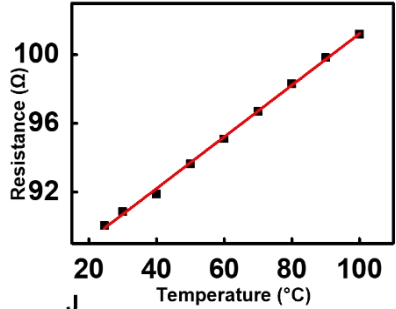
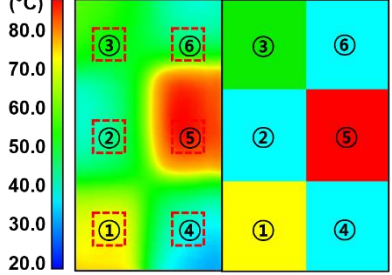
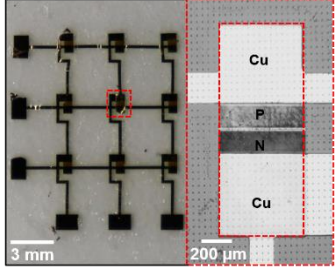
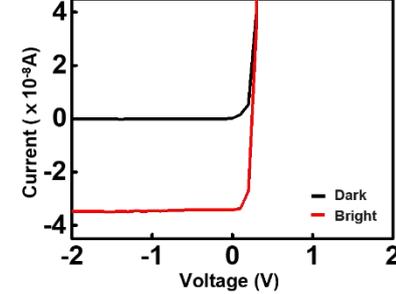
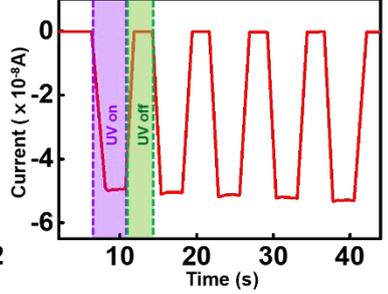

**Fig. 5. Direct integration of multipurpose sensors onto the transient soft actuator.** (**A**) Temperature and strain sensors integrated into the soft actuator (left); exploded view of both sensors (right). (**B**) Photodetector array integrated into the soft actuator (left); exploded view of the photodetector array (right). (**C**) Cu strain sensor with 200 µm-wide electrode traces. (**D**) Variation in the strain sensor capacitance as the bending angle increases (0° to 35°). (**E**) Capacitance change in the strain sensor under cyclical pneumatic pressurization between -10 kPa and +10 kPa. (**F**) Cu temperature sensor with 200 µm-wide electrode traces. (**G**) Resistance variation of the Cu temperature sensor as the ambient temperature increases. A coefficient of resistance of 0.2 $\Omega \cdot °C^{-1}$ for 25–100°C was measured from the plot. (**H**) Localized temperature variations of Cu resistive sensors with Joule-heated rods placed directly below (sensor 1:50°C, sensor 3:70°C, sensor 5:100°C) measured by an infrared camera (left) and a resistive temperature sensor array (right). (**I**) A 3 × 3 array of PIN photodiodes comprising a silicon membrane with a 20 µm channel length and 625 µm width, and Cu electrodes. (**J**) Current-voltage (I-V) characteristics of a photodetector under dark (light off, black) and bright (365 nm UV light on, red) conditions. (**K**) Photodiodic behavior during cycling UV on and off states. A current of $-5 \times 10^{-8}$ A is induced at -2 V and no current flows at 0 V.

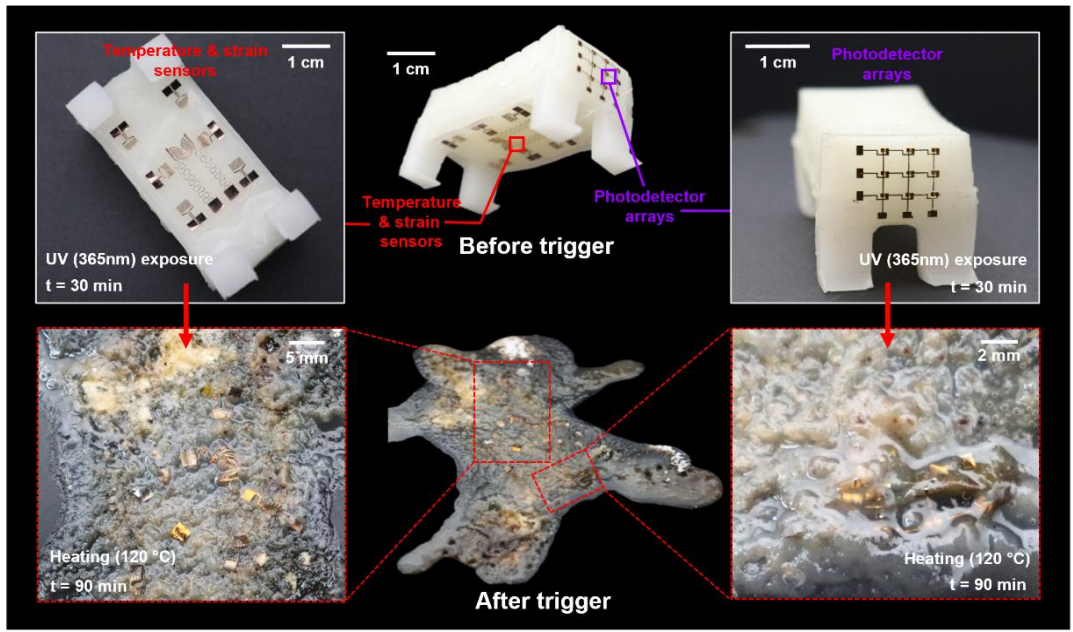

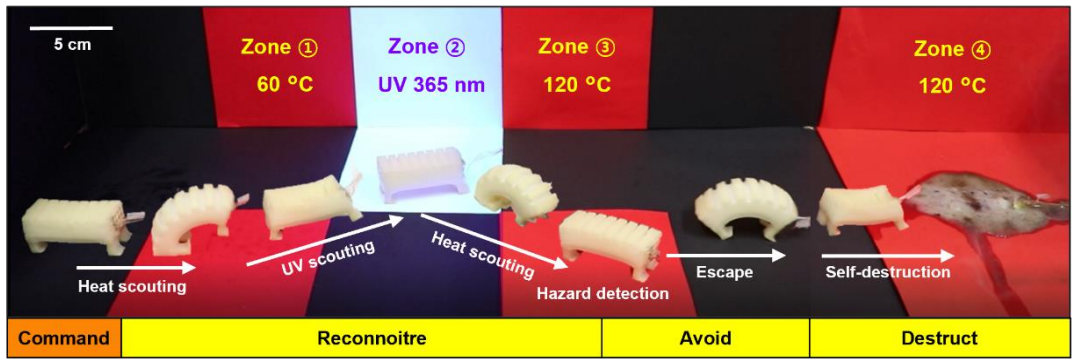

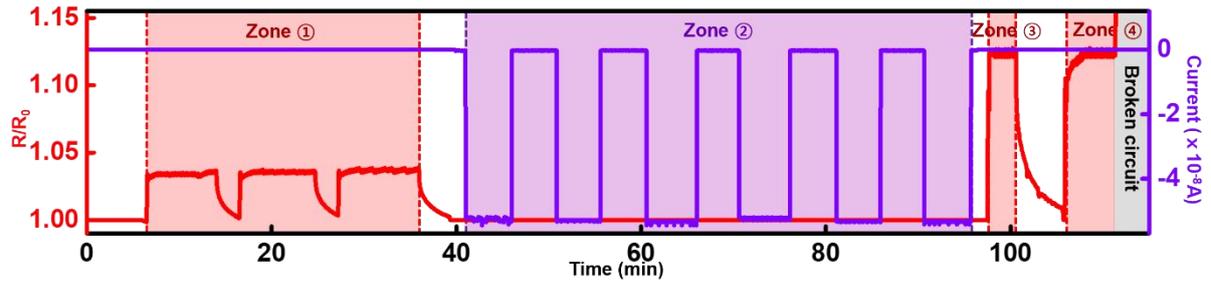

**Fig. 6. Autonomic operation of a life-time controlled gaiting robot.** (**A**) Photographs of the gaiting robot prior to and after decomposition at 120°C triggered by 30 min of UV (365 nm) exposure. (**B**) Time-lapse images of the gaiting robot with integrated electronic devices. The robot enters the heat zone (Zone 1, 70°C) and UV trigger zone (Zone 2, 365 nm) to gather temperature and UV light information. When the robot enters the high-temperature zone (Zone 3, 120°C), the embedded temperature sensor allows the robot to recognize accelerated decomposition conditions and escape from the hazardous zone. Once the robot decides to terminate its existence, it enters the final heat zone (Zone 4, 120°C) to fully disintegrate. (**C**) In situ measurement of temperature and UV light from the robot via embedded temperature sensor and photodetector. The conditions of the temperature and UV light correspond to the zones that the robot explored.

# Supplementary Materials

Supplementary Text 1: Kinetics of the photolysis of DPI-HFP

Supplementary Text 2: Details of the Vyazovkin's model

Fig. S1. Chemical structure analysis of DPI-HFP/Sylgard-184.

Fig. S2. $^{29}$Si NMR analyses of decomposition residues.

Fig. S3. Component analyses of DPI-HFP/PDMS decomposition residue.

Fig. S4. Photo-triggered decomposition of a silicone elastomer by a photo-fluoride generator.

Fig. S5. DSC analysis of DPI-HFP/Ecoflex 00-30 at various temperatures.

Fig. S6. Kinetic analysis of DPI-HFP/Sylgard-184.

Fig. S7. Decomposition of silicone elastomer composites by a photo-fluoride generator.

Fig. S8. Mechanical properties of DPI-HFP/Sylgard-184.

Fig. S9. Tensile testing of Ecoflex composites with various DPI-HFP concentrations.

Fig. S10. Dimension schematics of the pneumatic walking robot.

Fig. S11. FEM-analyzed strain distributions of the gaiting robot.

Fig. S12. Schematic overview of temperature and strain sensor fabrication.

Fig. S13. Conversion of capacitance to strain.

Fig. S14. Schematic overview of photodiode array fabrication.

Fig. S15. Transient behaviors of electronics integrated with the decomposing pneumatic soft robot.

Movie S1. Hypothetical military operation of the on-demand transient soft robot.

**Supplementary Text 1. Kinetics of the photolysis of DPI-HFP**

For a first-order kinetic reaction of DPI-HFP → HF,

$$v = k[DPI\text{-}HFP] = -\frac{d[DPI\text{-}HFP]}{dt} = \frac{d[HF]}{dt}$$

where $v$ is the rate of reaction, $k$ is the rate constant of DPI-HFP photolysis, $t$ is the time, and [DPI-HFP] is the concentration at this time point. By arranging the above equation based on the variable [DPI-HFP], we get:

$$-kdt = \frac{d[DPI\text{-}HFP]}{[DPI\text{-}HFP]}$$

$$[DPI\text{-}HFP] = [DPI\text{-}HFP]_0 e^{-kt}$$

where $[DPI\text{-}HFP]_0$ is the initial concentration. Considering one DPI-HFP yields a stoichiometric amount of HF, the equation can be rearranged as:

$$[DPI\text{-}HFP] = [DPI\text{-}HFP]_0 e^{-kt} = [DPI\text{-}HFP]_0 - [HF]$$

$$[HF] = [DPI\text{-}HFP]_0 (1 - e^{-t})$$

**Supplementary Text 2. Details of Vyazovkin's model**

In Vyazovkin's model, the kinetics of the solid-to-liquid transition are expressed as:

$$\frac{d\alpha}{dt} = k(T) \times f(\alpha) \tag{S1}$$

where α is the extent of phase conversion, k(T) is the rate constant as a function of temperature, and f(α) is a function dependent on the extent of phase conversion. Because Si—O cleavage occurs with a finite amount of fluoride and is a decelerating reaction, f(α) can be expressed for the decomposition reaction as:

$$f(\alpha) = (1-\alpha)^n \tag{S2}$$

where *n* is the reaction order. As the reaction order of Si—O cleavage with HF is 1, $f(\alpha) = (1-\alpha)$ in this case.

Furthermore, combining Eqs. (1) and (2) from the main text yields Eq. (S3). Differentiating Eq. (S3) by variable *t* leads to Eq. (S4),

$$\Delta H_t = \Delta H_{total}(1 - e^{-kt}) \tag{S3}$$

$$\frac{d(\Delta H_t)}{dt} = k\Delta H_{total} \times e^{-kt} \tag{S4}$$

Because the isothermal photo-DSC curve is a plot of $\frac{d(\Delta H_t)}{dt}$ over time *t*, fitting the DSC curve with Eq. (S4) yields the *k* value. Fig. 3 shows the photo-DSC curve and *k* values obtained for DPI-HFP/Ecoflex 00-30 at varying temperatures.

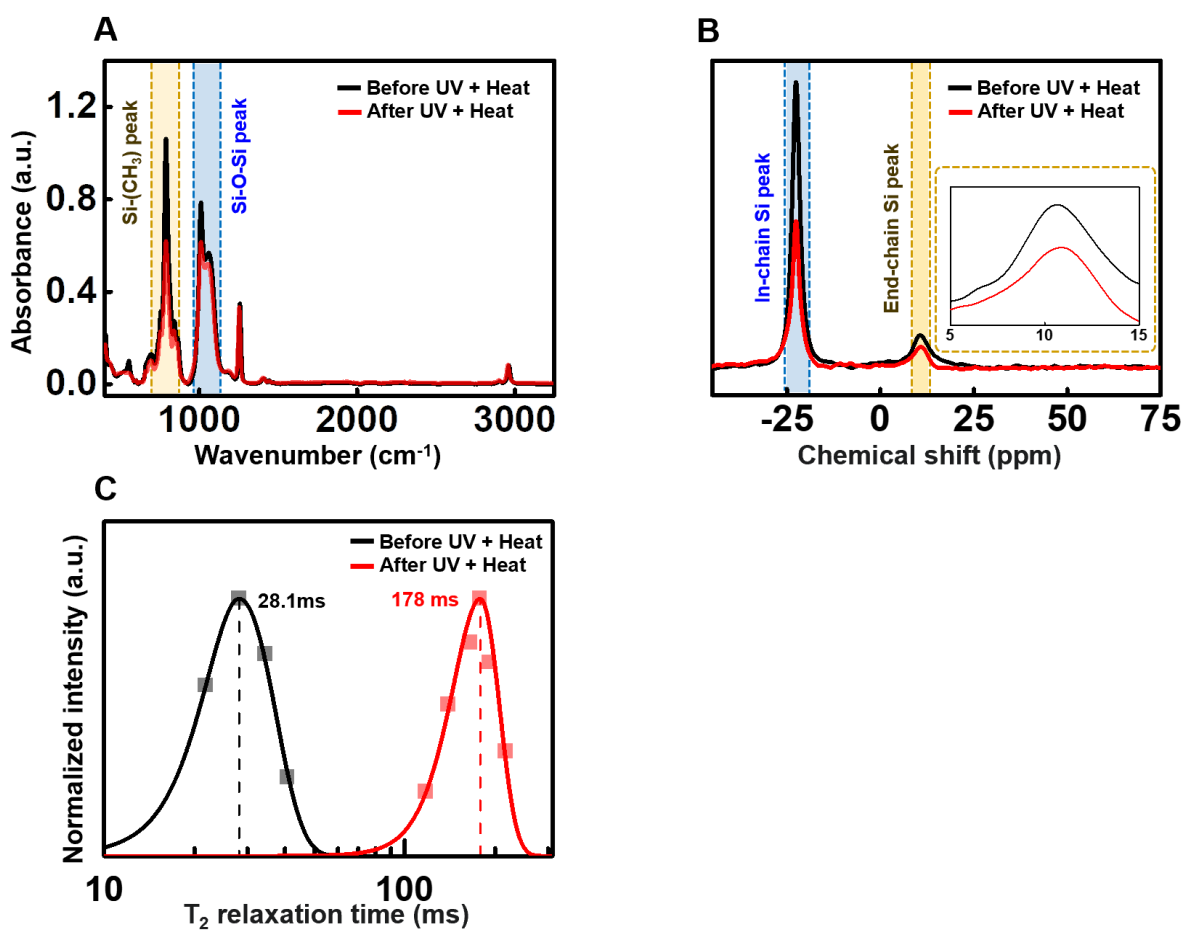

**Fig. S1. Chemical structure analysis of DPI-HFP/Sylgard-184.** (**A**) FT-IR spectra of DPI-HFP/Sylgard-184, showing a decrease in the absorbance in the Si–CH$_3$ peak (792 nm$^{-1}$) and Si–O–Si peak (1013 nm$^{-1}$) after UV and heat triggers. (**B**) Solid-state $^{29}$Si NMR spectra of DPI-HFP/Sylgard-184, showing a decrease in the intensity of the Si–CH$_3$ peak (-22.56 ppm) and Si–O–Si peak (6.56 ppm) after UV and heat exposure. (**C**) Solid-state $^1$H NMR Transverse relaxation time (T$_2$) distributions of DPI-HFP/Sylgard-184. The T$_2$ relaxation time increased after UV and heat exposure, indicating a decrease in the crosslinking density.

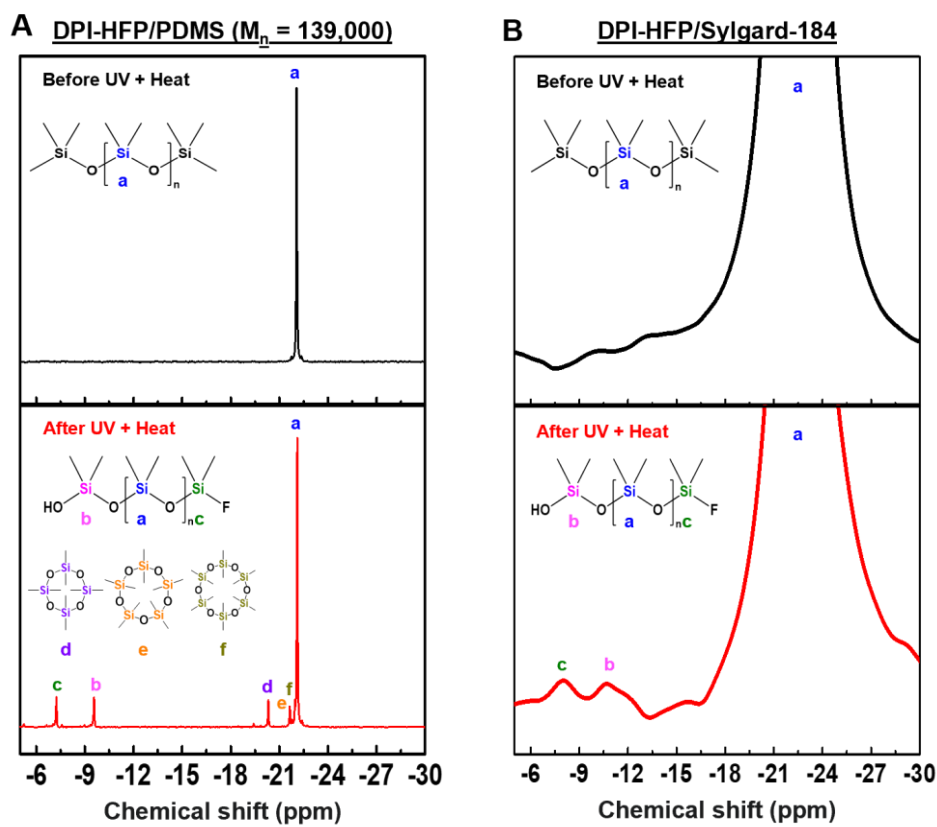

**Fig. S2. $^{29}$Si NMR analyses of decomposition residues.** (**A**) $^{29}$Si NMR spectra of DPI-HFP/PDMS before (top) and after (bottom) decomposition is triggered by UV light and heat, demonstrating the formation of cyclic $D_4$, $D_5$, and $D_6$ siloxane compounds and hydroxyl- and fluoride-terminated PDMS chains after applying the trigger. (**B**) Solid-state $^{29}$Si NMR spectra of DPI-HFP/Sylgard-184 before (top) and after (bottom) decomposition is triggered by UV light and heat, showing results identical to those of DPI-HFP/Ecoflex 00-30.

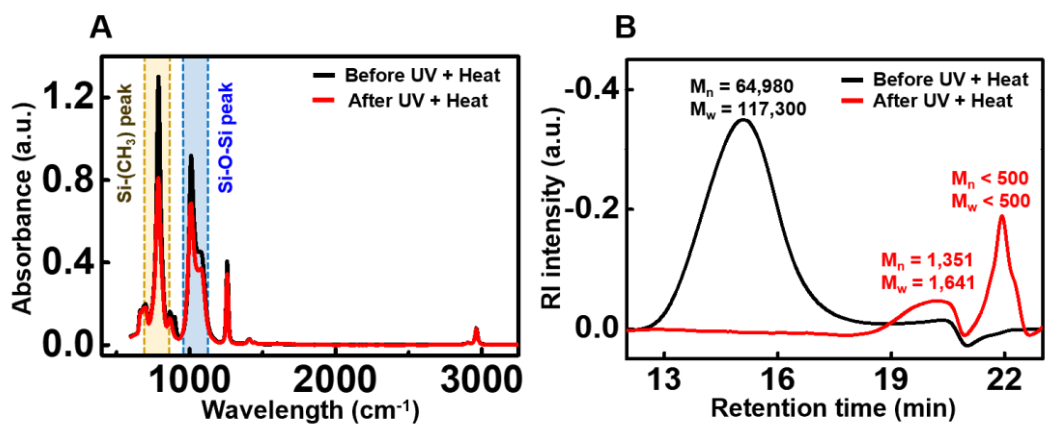

**Fig. S3. Component analyses of DPI-HFP/PDMS decomposition residue.** (**A**) FT-IR spectra of DPI-HFP/PDMS, showing decreased absorbance in the Si–CH$_3$ peak (786 cm$^{-1}$) and Si–O–Si peak (1014 cm$^{-1}$) after UV and heat treatment. (**B**) Gel permeation chromatography results of DPI-HFP/PDMS, showing decreased molecular weight after UV and heat treatment.

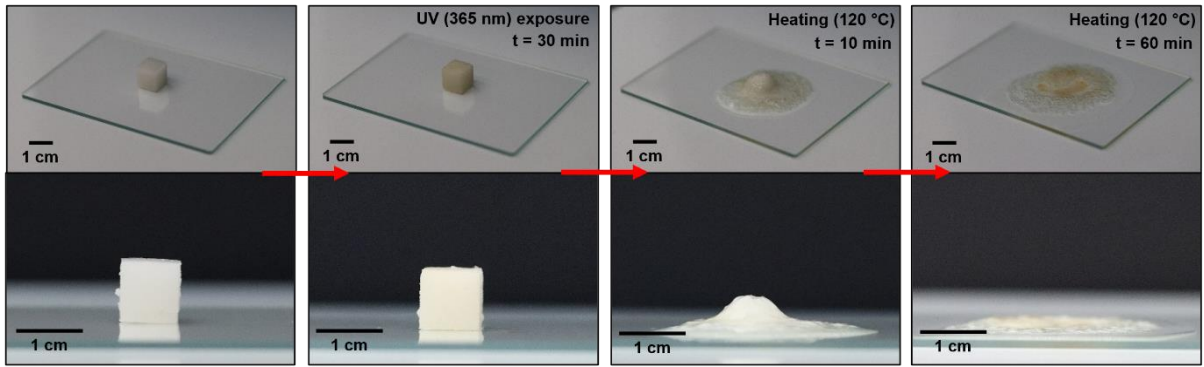

**Fig. S4. Photo-triggered decomposition of a silicone elastomer by a photo-fluoride generator.** Decomposition of DPI-HFP/Sylgard-184 at 120°C over 1 h after 30 min 365 nm UV exposure.

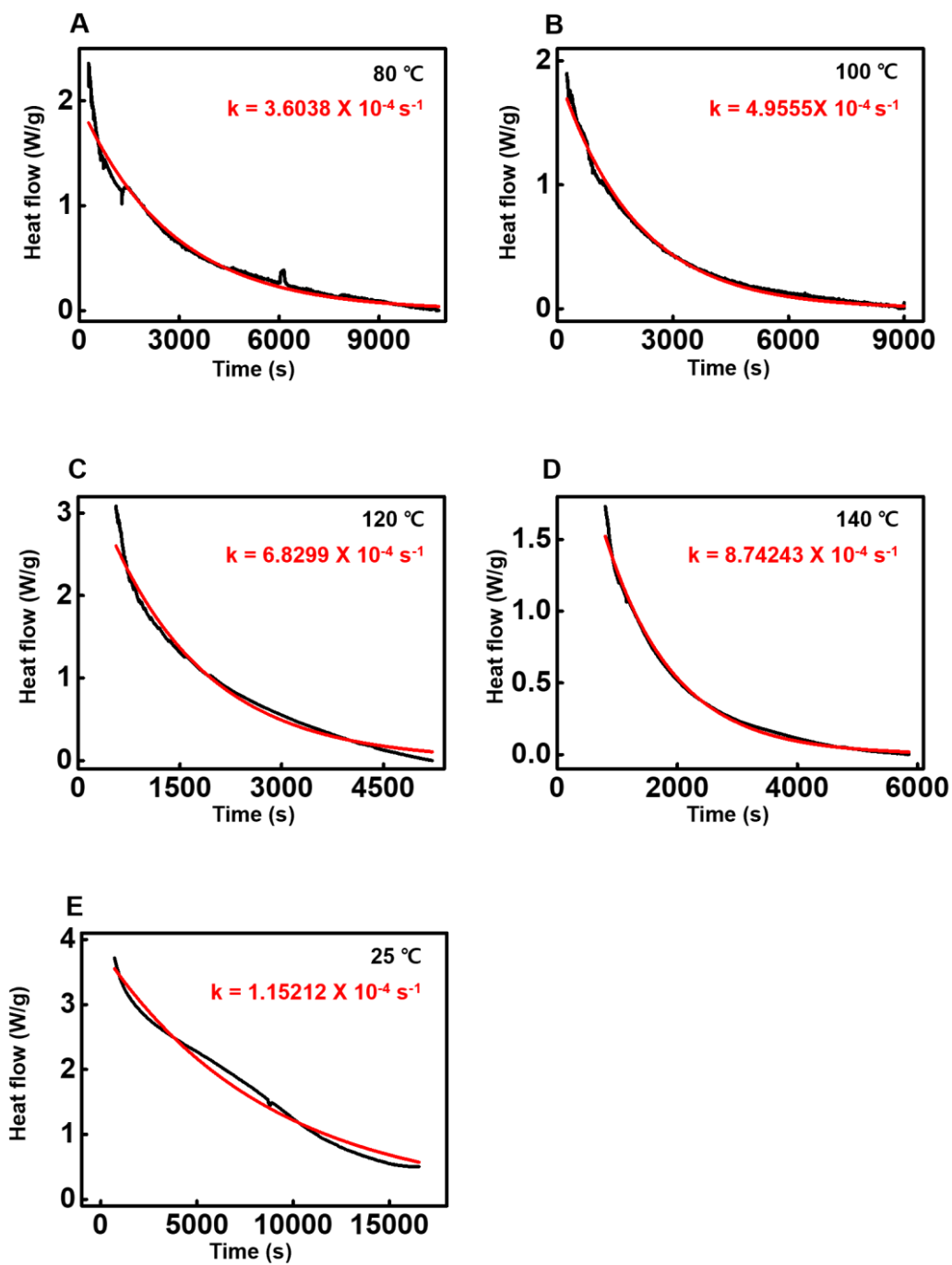

**Fig. S5. DSC analysis of DPI-HFP/Ecoflex 00-30 at various temperatures.** (**A-E**) DSC curves of DPI-HFP/Ecoflex 00-30 with UV radiation at different temperatures. (**A**) 80°C, (**B**) 100°C, (**C**) 120°C, (**D**) 140°C, and (**E**) 25°C.

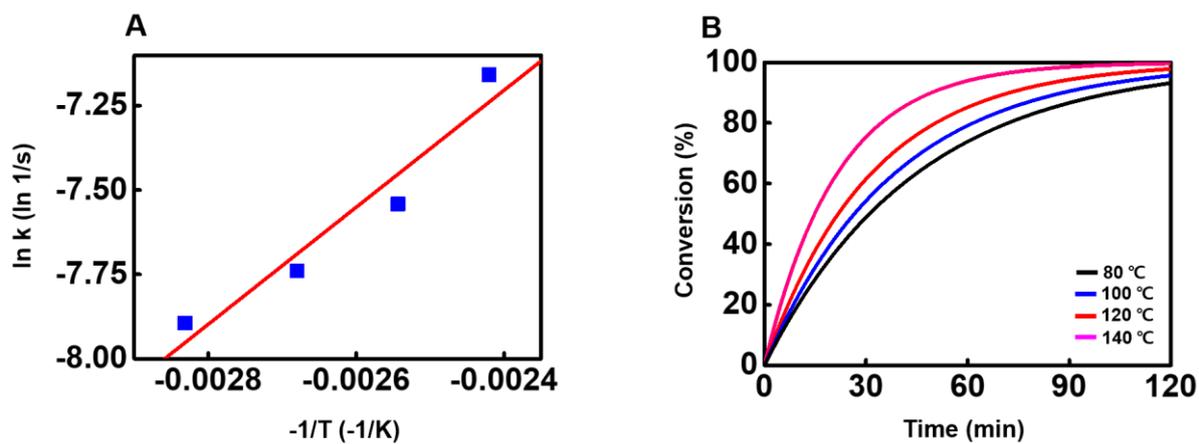

**Fig. S6. Kinetic analysis of DPI-HFP/Sylgard-184.** (**A**) Arrhenius plot of DPI-HFP/Sylgard-184 used for calculating the kinetic parameters of the UV triggered decomposition reaction. (**B**) Phase conversion percentage of DPI-HFP/Sylgard-184 with UV radiation at different temperatures (80°C, black; 100°C, blue; 120°C, red; 140°C, magenta).

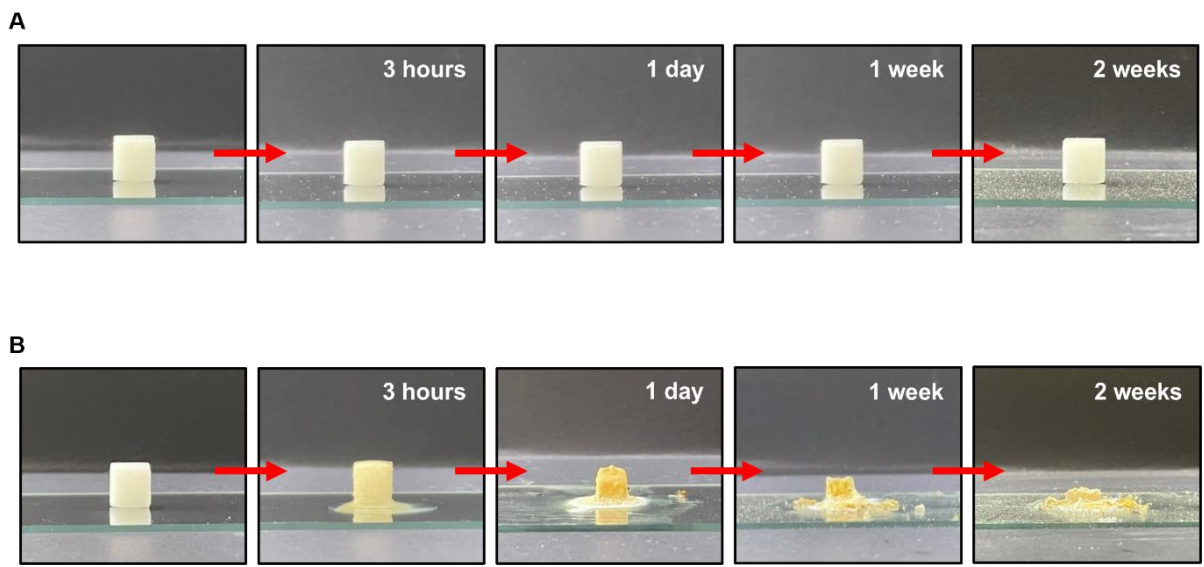

**Fig. S7. Decomposition of silicone elastomer composites by a photo-fluoride generator.** (**A**) Lack of decomposition of DPI-HFP/Ecoflex 00-30 at room temperature under daylight exposure. (**B**) Decomposition of DPI-HFP/Ecoflex 00-30 at room temperature under 365 nm UV exposure.

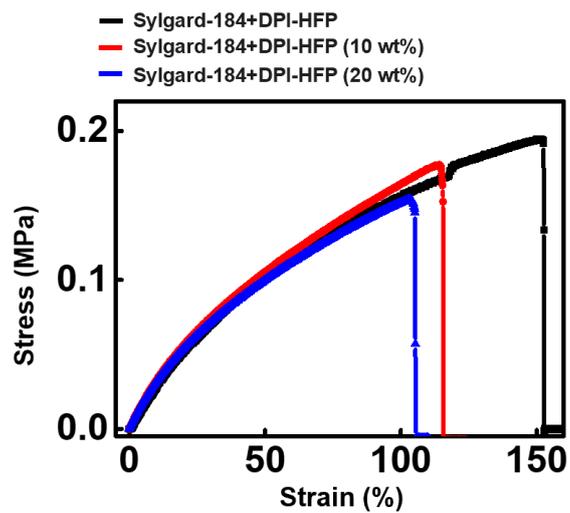

**Fig. S8. Mechanical properties of DPI-HFP/Sylgard-184.** Stress–strain behavior of DPI-HFP/Sylgard-184 at various DPI-HFP concentrations (0 wt%, black; 10 wt%, red; 20 wt%, blue).

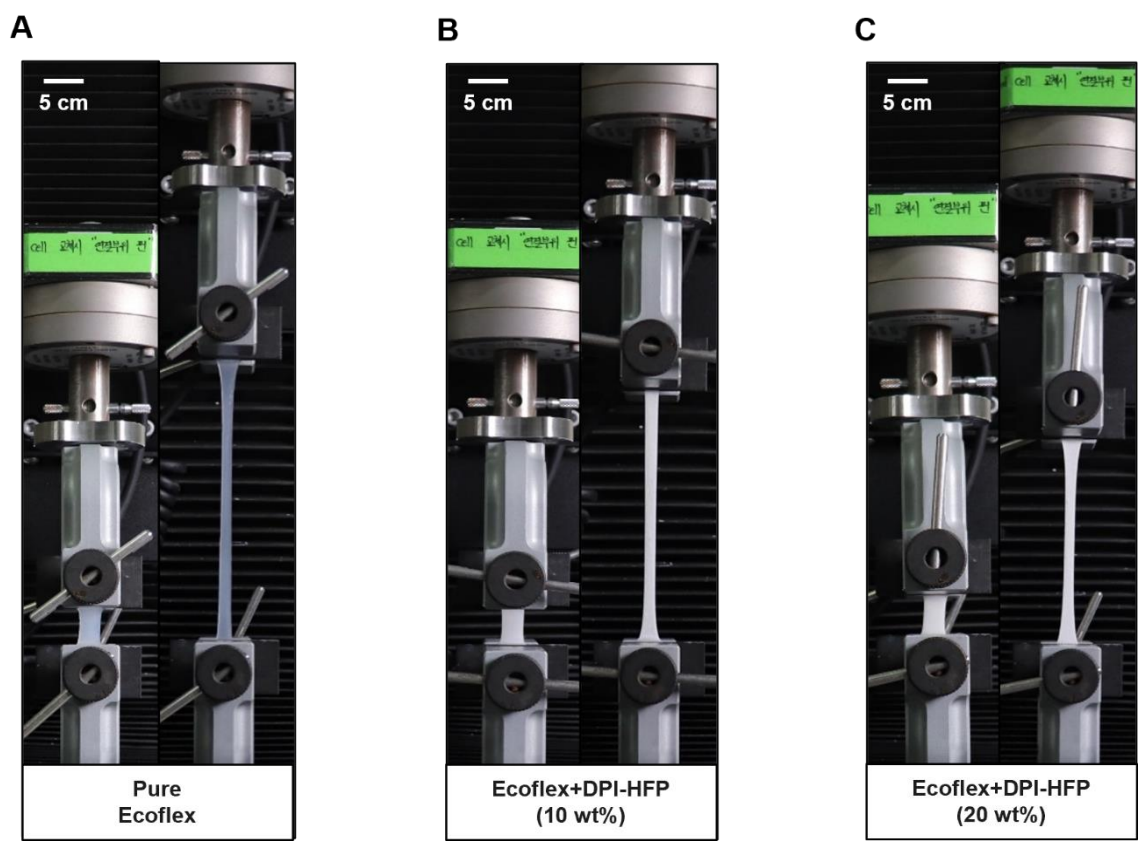

**Fig. S9. Tensile testing of Ecoflex composites with various DPI-HFP concentrations.** (**A-C**) Photos of the elongation of DPI-HFP/Ecoflex 00-30 at (**A**) 0 wt%, (**B**) 10 wt%, and (**C**) 20 wt% concentration of DPI-HFP.

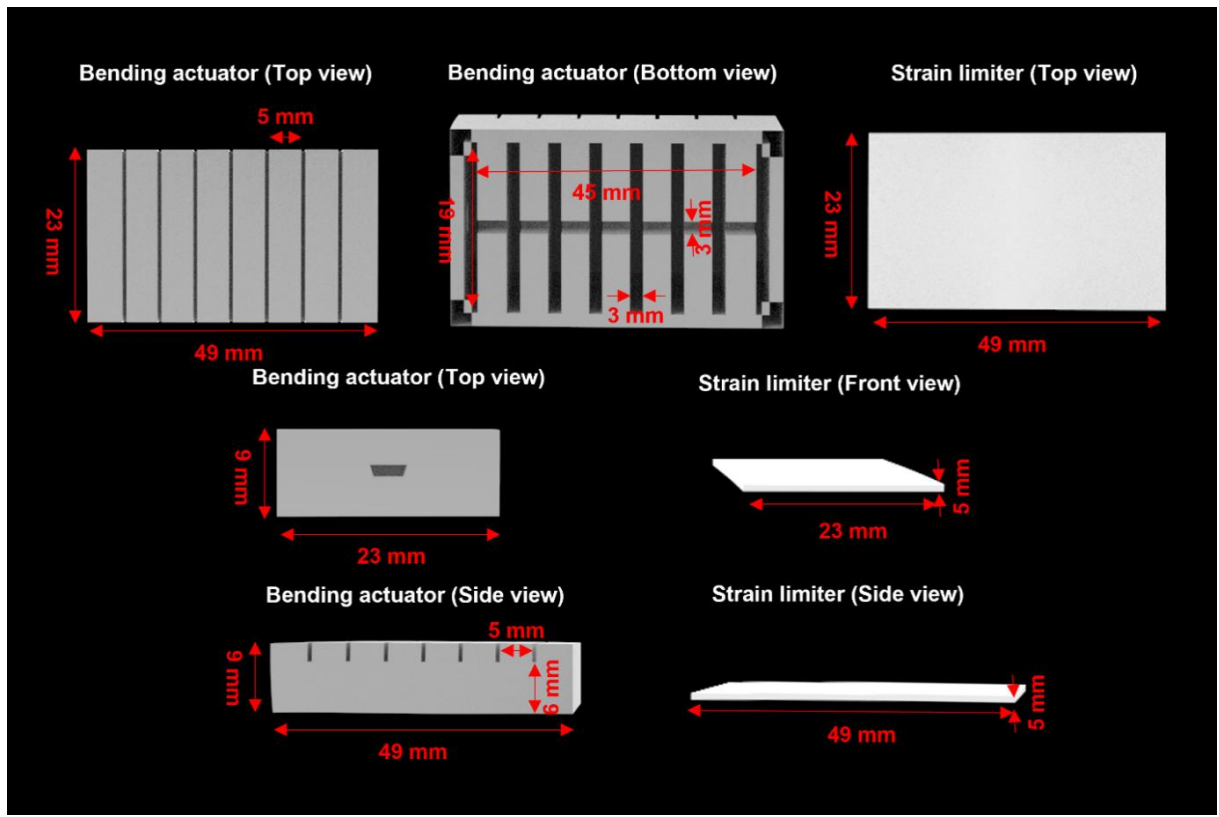

**Fig. S10. Dimension schematics of the pneumatic walking robot.** Digital renderings of the bending actuator and strain limiter of the pneumatically actuated walking soft robot labeled with their dimensions.

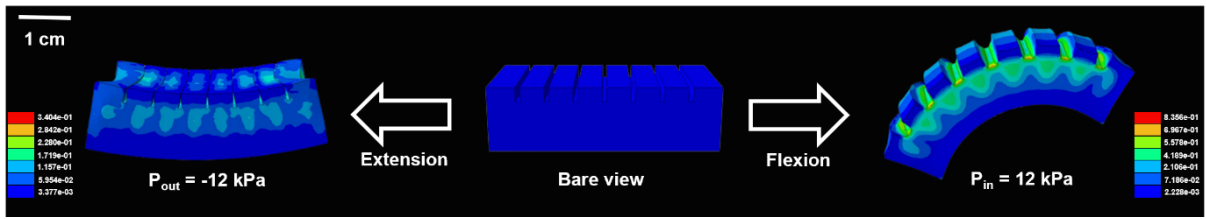

**Fig. S11. FEM-analyzed strain distributions of the gaiting robot.** The FEM image shows the external stress concentrations of the robot during simulated extension (left) and flexion (right) through pneumatic actuation.

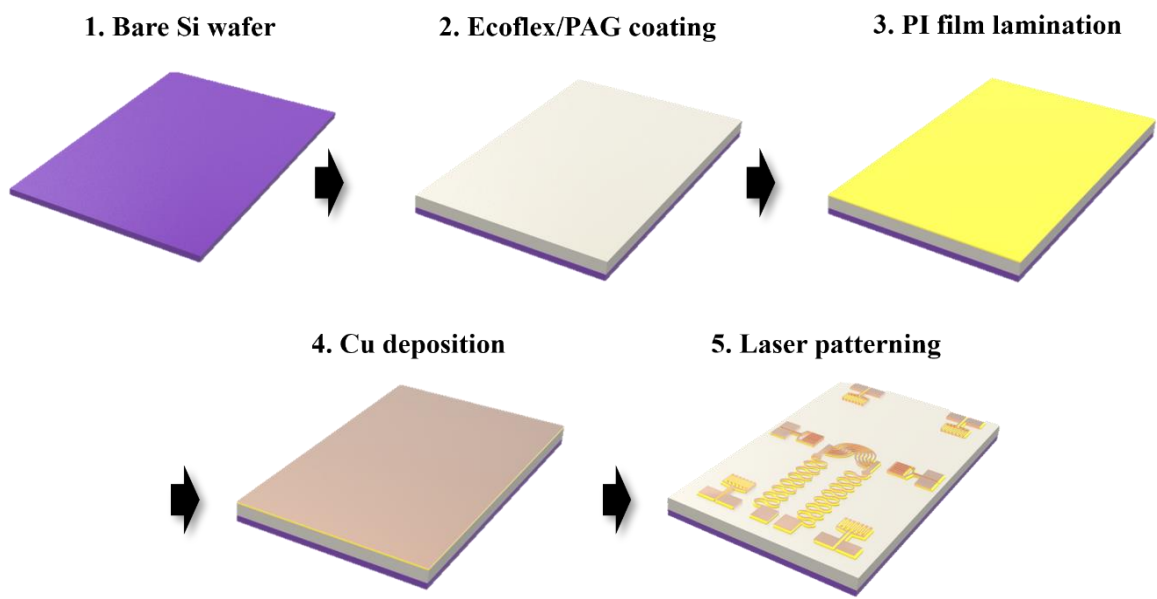

**Fig. S12. Schematic overview of temperature and strain sensor fabrication.**

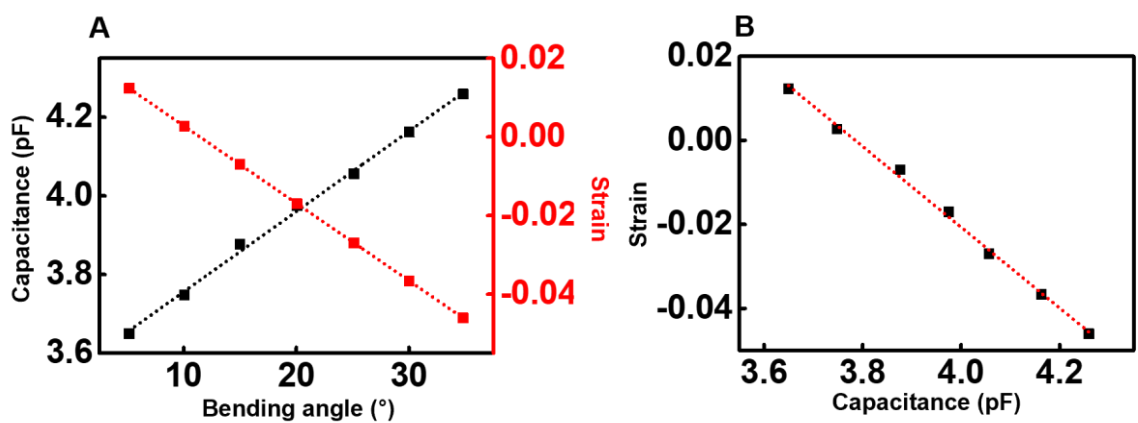

**Fig. S13. Conversion of capacitance to strain.** (**A**) Bending angle-dependent measurements of strain sensor capacitance (experimental, black) and actuator strain (FEM analysis, red). (**B**) Strain sensor capacitance plotted as a function of strain using the data from

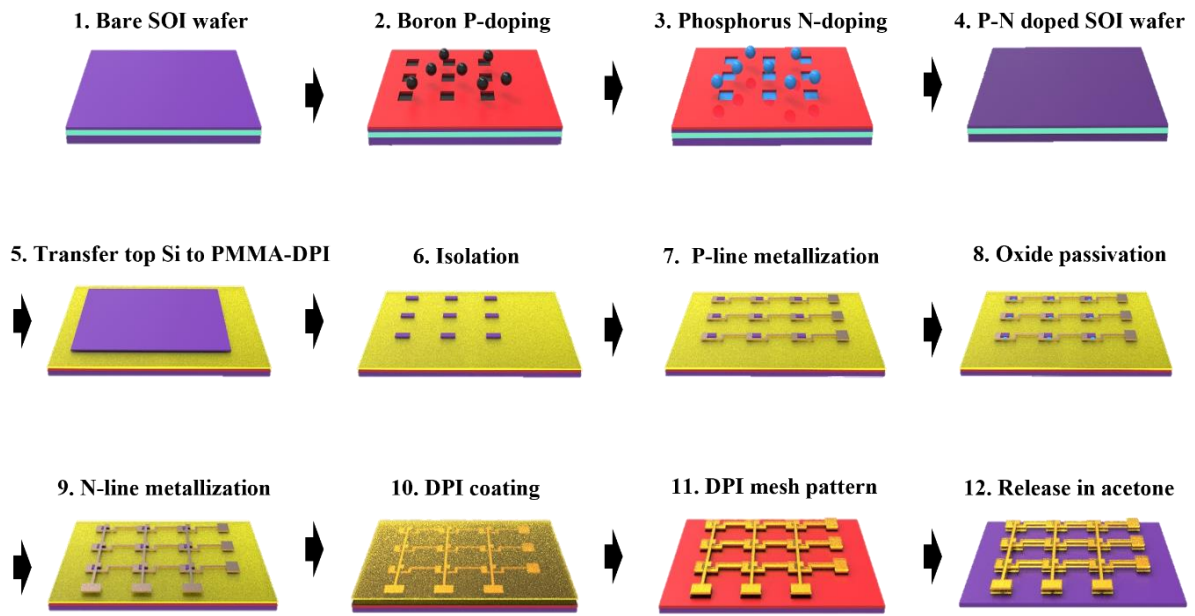

**Fig. S14. Schematic overview of photodiode array fabrication.**

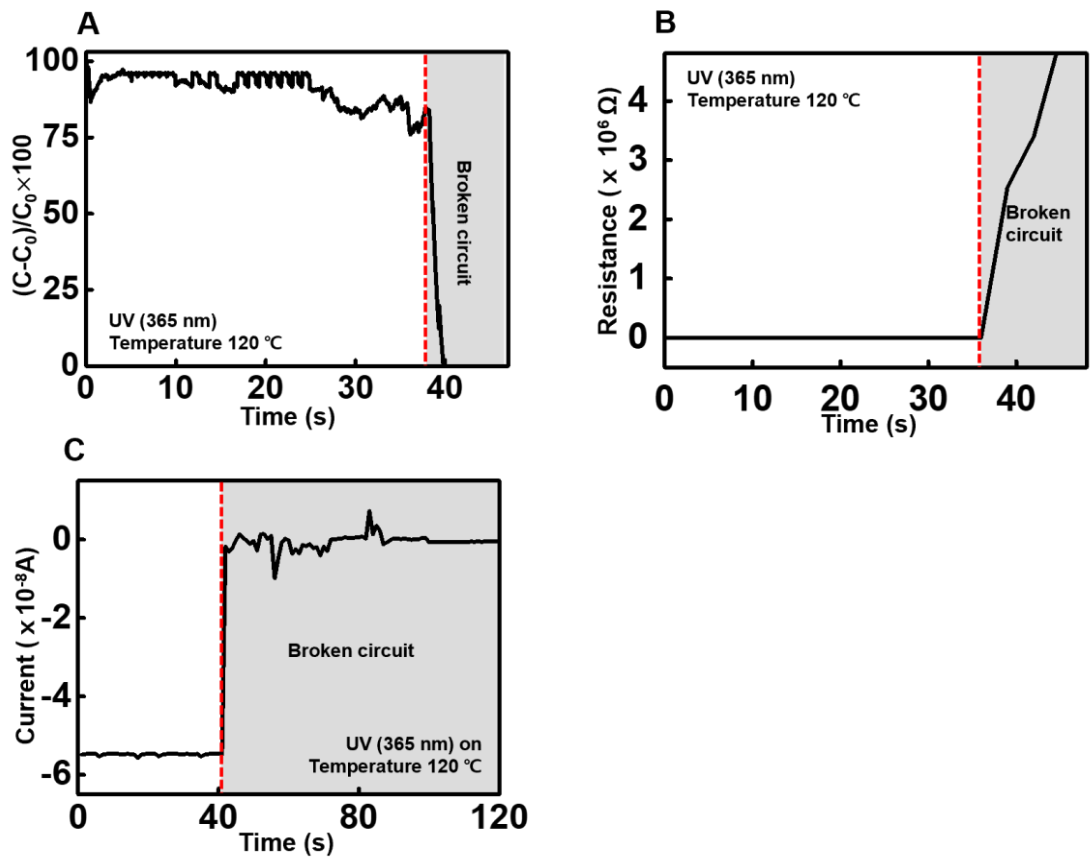

**Fig. S15. Transient behaviors of electronics integrated with the decomposing pneumatic soft robot.** (**A**) Time-dependent changes in capacitance measured by the strain sensor. Transient behavior is observed when the capacitance reading is disabled. (**B**) Time-dependent changes in resistance measured by the temperature sensor. Transient behavior is observed when the resistance rapidly increases to $10^6$ Ω. (**C**) Time-dependent changes in the photocurrent measured by the photodetector array. Transient behavior is observed when the induced photocurrent reaches 0 A.